\documentclass[10pt,twocolumn,letterpaper]{article}

\usepackage[pagenumbers]{cvpr} % To force page numbers, e.g. for an arXiv version
\makeatletter
\@namedef{ver@everyshi.sty}{}
\makeatother
\usepackage{tikz}

\usepackage{graphicx}
\usepackage{amsmath}
\usepackage{amssymb}
\usepackage{mathrsfs}
\usepackage{booktabs}
\usepackage{latexsym}
\usepackage{epsfig}
\usepackage{bm}
\usepackage{bbm} % \mathbbm
\usepackage{makecell,multirow,diagbox}
\usepackage{amsfonts}
\usepackage{colortbl}
\usepackage{microtype}
\usepackage{multirow, booktabs}
\usepackage{adjustbox}
\usepackage{enumitem}
\usepackage{float}
\usepackage{algorithm}
\usepackage{algorithmicx}
\usepackage{xcolor}
\usepackage{tcolorbox}
\usepackage{wasysym}
\usepackage{caption}
\usepackage{tcolorbox}

\usepackage{algorithm}
\usepackage{algpseudocode}
\usepackage{appendix}

\newcommand{\approach}{\textsc{BALLAD}}
\usepackage[pagebackref,breaklinks,colorlinks]{hyperref}

\usepackage[capitalize]{cleveref}
\crefname{section}{Sec.}{Secs.}
\Crefname{section}{Section}{Sections}
\Crefname{table}{Table}{Tables}
\crefname{table}{Tab.}{Tabs.}

\def\cvprPaperID{2828} % *** Enter the CVPR Paper ID here
\def\confName{CVPR}
\def\confYear{2022}

\begin{document}

%%%%%%%%% TITLE - PLEASE UPDATE
\title{A Simple Long-Tailed Recognition Baseline via Vision-Language Model}
% \title{BALLAD: A Simple Baseline for Long-Tailed Recognition}
\author{Teli Ma$^{1}$, Shijie Geng$^{1}$, Mengmeng Wang$^{1}$, Jing Shao$^{2}$, Jiasen Lu$^{3}$ \\ Hongsheng Li$^{4}$,  Peng Gao$^{\dagger 1}$, Yu Qiao$^1$\\ 
  $^1$Shanghai Artificial Intelligence Laboratory \quad 
  $^2$SenseTime Research \\
  $^3$PRIOR@Allen Institute for AI \quad
  $^4$The Chinese University of Hong Kong  \\
\texttt{\{mateli, gaopeng ,qiaoyu\}@pjlab.org.cn}
}

\maketitle

% Ballads derive from the medieval French chanson balladée or ballade, which were originally "dance songs". 
% The harmony of head and tail classes

% long-tailed distribution of open classes -> 1) few-shot; 2) imbalance; 3) open classes
% VL-Model: 3
% Adapter: 1
% Two-Stage: 2 (Stage 1: for head class, Stage 2: for tail class)

%%%%%%%%% ABSTRACT
\begin{abstract}
The visual world naturally exhibits a long-tailed distribution of open classes, which poses great challenges to modern visual systems. Existing approaches either perform class re-balancing strategies or directly improve network modules to address the problem. However, they still train models with a finite set of predefined labels, limiting their supervision information and restricting their transferability to novel instances. Recent advances in large-scale contrastive visual-language pretraining shed light on a new pathway for visual recognition. With open-vocabulary supervisions, pretrained contrastive vision-language models learn powerful multimodal representations that are promising to handle data deficiency and unseen concepts. By calculating the semantic similarity between visual and text inputs, visual recognition is converted to a vision-language matching problem. Inspired by this, we propose \textbf{BALLAD} to leverage contrastive vision-language models for long-tailed recognition. We first continue pretraining the vision-language backbone through contrastive learning on a specific long-tailed target dataset. Afterward, we freeze the backbone and further employ an additional adapter layer to enhance the representations of tail classes on balanced training samples built with re-sampling strategies. Extensive experiments have been conducted on three popular long-tailed recognition benchmarks. As a result, our simple and effective approach sets the new state-of-the-art performances and outperforms competitive baselines with a large margin. Code is released at \url{https://github.com/gaopengcuhk/BALLAD}.
% (BALanced Linear ADapter) 
\end{abstract}
% , namely ImageNet-LT and Places-LT. 
%%  Deep Neural Networks offered by faster GPU
%%  https://en.wikipedia.org/wiki/Long_tail
%%  Skewness -> 偏态（Skewness）：用于描述数据分布左右对称性的指标
%%  背景 长尾: Zipf & Pareto 
%%  Depicted by ...
%%%%%%%%% BODY TEXT

\begin{figure}[ht!]
\centering
\includegraphics[width=0.96\linewidth,trim={0cm 0cm 0cm 0cm}]{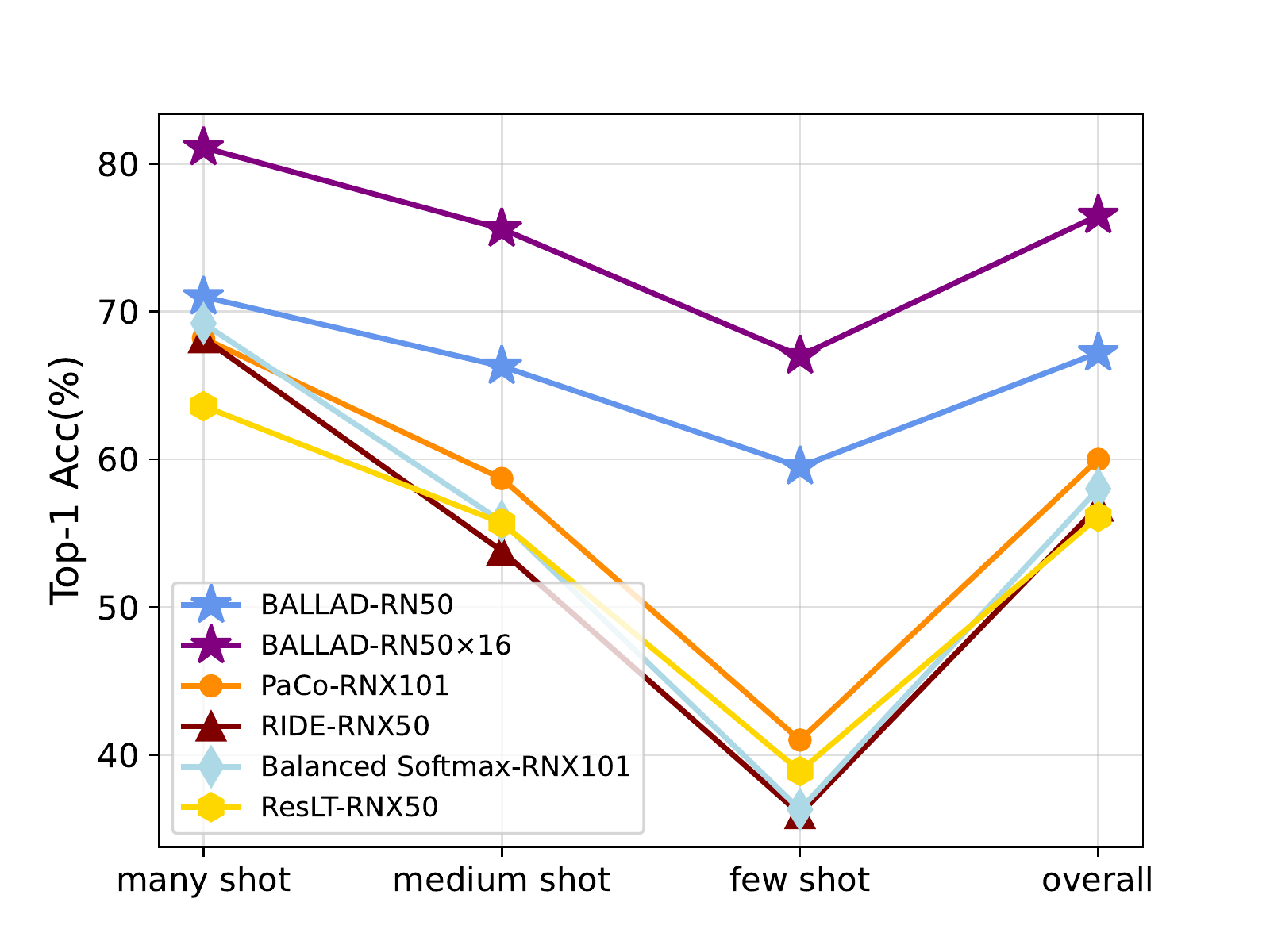}
\vspace*{-13pt}
\caption{Comparison with the state-of-the-art approaches on ImageNet-LT. \approach{} with ResNet-50 visual backbone outperforms state-of-the-art models which have more complex backbones and longer training epochs by a large margin (up to \textbf{+7.2\%} improvement on overall accuracy compared with PaCo~\cite{cui2021parametric}), especially for categories without abundant training examples.~\approach{} with ResNet-50$\times$16 achieves a new state-of-the-art performance of \textbf{76.5\%} top-1 accuracy on ImageNet-LT benchmark dataset. (RN50: ResNet-50, RN50$\times$16: the EfficientNet method with 16$\times$ compute of ResNet-50, RNX50: ResNeXt-50, RNX101: ResNeXt-101.)}
\vspace*{-13pt}
\label{fig:intro}
\end{figure}

\section{Introduction}
\label{sec:intro}
% 引出Data不平衡
During past years, visual recognition tasks, such as image classification~\cite{simonyan15,he2016deep,xie2017aggregated}, object detection~\cite{ren2015faster,lin2017feature}, semantic segmentation~\cite{long2015fully,chen2017deeplab,zhao2017pyramid}, and instance segmentation~\cite{he2017mask,hu2018learning,liu2018path} have been significantly improved. 
%The performance gains can be attributed to the availability of faster GPUs, large-scale high-quality datasets~\cite{deng2009imagenet,lin2014microsoft,krishna2017visual}, carefully-designed training techniques~\cite{wu2018group,he2020momentum,tan2021efficientnetv2}, and better neural network architectures~\cite{carion2020end,dosovitskiy2021an,Touvron_2021_ICCV,Liu_2021_ICCV,gao2021container}.
The performance gains can be largely attributed to the availability of large-scale high-quality datasets~\cite{deng2009imagenet,lin2014microsoft,krishna2017visual}.
%The unceasing increase of computation resources and deep network capacities demands more data from more categories to be collected.
However, the problem of data imbalance has inevitably emerged since real-world data often abide by a long-tailed distribution (\eg, Pareto distribution\cite{pareto1964cours} or Zipf's law~\cite{zipf1949human}). In other words, a few head classes dominate the majority of training examples, whereas many rare or fine-grained classes only have limited relevant data points.

% 引出long-tailed recognition, 
To alleviate the issue, previous efforts either carefully create more balanced datasets (\eg, ImageNet~\cite{deng2009imagenet}, MSCOCO~\cite{lin2014microsoft}, and Kinetics-400~\cite{kay2017kinetics}) with human labors or develop more robust algorithms to handle data imbalance. However, since the former is notoriously laborious and expensive, many researchers have been devoted to the latter. Formally, long-tailed recognition (LTR) is a research field seeking robust models that 1) are resistant to significant imbalanced class distribution; 2) can deal with few-shot learning of tail classes.
% 3) has potential for generalization to unseen novel categories.
% \mateli{3 is abrupt since the generalization is not common in previous work.}
Many methods~\cite{zhang2021deep} have been proposed for solving LTR problems. According to the core technical contributions, they can be divided into two categories. Methods in the first line focus on class re-balancing strategies~\cite{menon2021longtail,hong2021disentangling,zhang2021distribution,kang2019few} such as data re-sampling, loss re-weighting, and logit adjustment. The second category focuses on improving network modules~\cite{cui2021parametric,Kang2020Decoupling,zhang2021test,Samuel_2021_ICCV,tang2020long,cui2021reslt,zhou2020bbn} by classifier designing, decoupled training, and representation learning. 
While these methods have achieved significant progress, the performance of LTR remains unsatisfactory. When delving deeper into the utilization of the existing imbalance datasets, we have observed that almost all previous efforts are confined to a predetermined manner which designs models entirely relying on the visual modality. That is to say, they totally ignore the semantic features of the raw label text, which may be a promising solution to exert additional supervision on inadequate data sources. 
% These approaches have made great progress in balancing the learning of different categories and keeping tail classes from being overwhelmed by dominant ones. 
% Nevertheless, almost all previous efforts are confined to a predetermined manner of designing methods entirely based on visual modality. They omit that learning from the raw label text may be a promising solution to exert additional supervision on inadequate data sources. 
%largely broadening generalization abilities to few-shot categories and zero-shot novel instances. 
Therefore, this paper explores whether language modality can be effective and complementary information for this task. In the meantime, we could broaden generalization abilities to few-shot categories and zero-shot novel instances.
%Nevertheless, they all train their models with a fixed set of discrete labels. This will largely constrain their generalization abilities to few-shot categories and zero-shot novel instances. Hence, they are far from meeting the above requirements sufficiently. 
% \jiasen{This claim is a little strong, and we didn't explain why all the prior method didn't meet the LTR requirements. Consider soften the claim or explain a little bit in details}.
% \shijie{I tried to soften the claim.}

% Those methods can be divided into two branches. Two-stage and Single-stage BLAH, BLAH. 
% carefully curated  high-quality balanced datasets like ImageNet, MSCOCO, and Kinetics-400 were  to facilitate the vision community, despite the construction process is costly and laborious. In contrast, designing robust deep models is obviously a better orientation to handle data imbalance.
% Construction of large-scale datasets is time-consuming and labor-intensive. 
% With the increase of available computation resources and modern neural network capacities, more data and more categories need to be collected for constructing larger datasets for efficient training of DNN. % Data imbalance problem will naturally appear when we build large scale datasets.
% Construction of carefully curated datasets like ImageNet, COCO and kinetics-400 is challenging for large-scale datasets. Thus, how to train a robust DNN given imbalanced data-sets is a key problem to be solved. 

\begin{figure*}[ht!]
\centering
\includegraphics[width=0.93\linewidth,trim={0cm 0cm 0cm 0cm}]{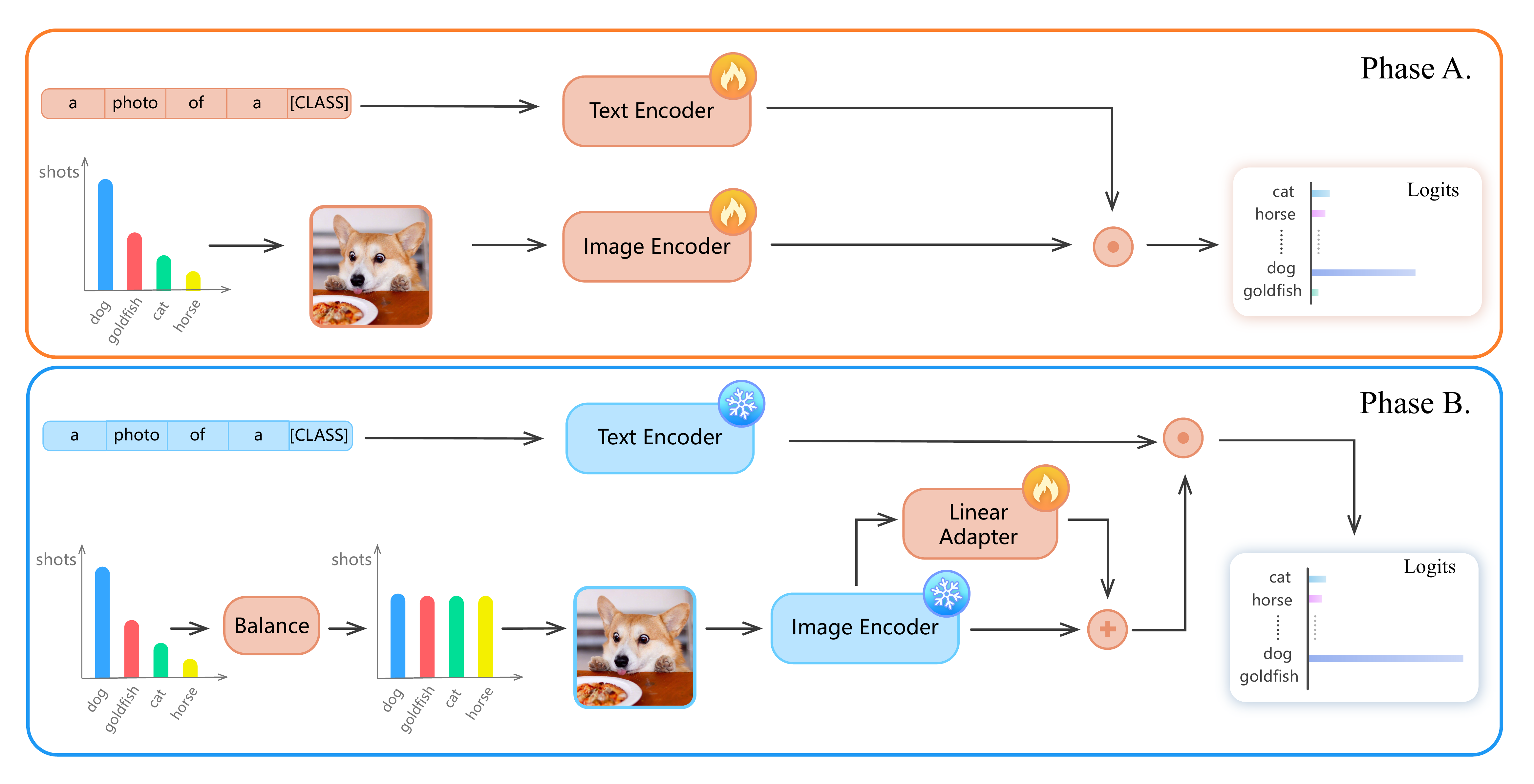}
\vspace*{-13pt}
\caption{Overview of our \approach{} framework. In Phase A, we keep pretraining the text and image branches of the vision-language backbone on long-tailed data. After Phase A, head classes typically achieve good enough classification performance, whereas tail classes are still far from perfect. During Phase B, a linear adapter is adopted to further train the vision-language backbone on balanced training samples. As a result, tail classes enjoy a performance boost while head classes slightly increase or maintain their original classification accuracy.
% The two phases are intended for handling head and tail classes respectively.
\includegraphics[height=8pt]{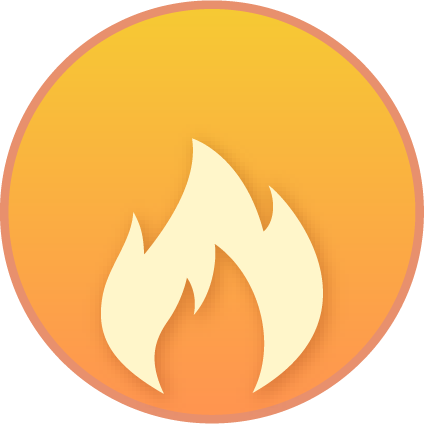} represents training with parameter update while \includegraphics[height=8pt]{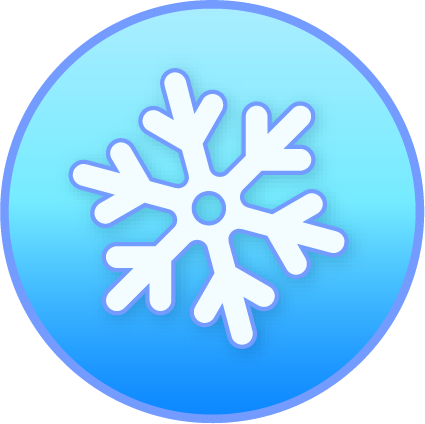} represents freezing parameters.}
\vspace*{-13pt}
\label{fig:1}
\end{figure*}

Recently, contrastive vision-language models such as CLIP~\cite{radford2021learning} and ALIGN~\cite{jia2021scaling} brought a breath of fresh air into the vision community. They learn to align vision and language representations with a contrastive loss given large-scale noisy image-text pairs collected from the web. The powerful visual-language representations obtained from pretraining significantly improve the zero-shot classification performance in open-vocabulary settings without any additional annotations.
% for VL-Model, learn representation -> equals -> refine feature + classifier
% very different from previous decoupled training approaches
% \shijie{2. VL-Model for LTR, for VL-Model, learning representation equals to refining both feature and classifier, which is very different from previous decoupled training approaches.}
% \shijie{3. Introduction of \approach{}.}
% achieves better results on many-shots due to the utilization of abundant annotations and low performance on few-shots due to imbalance in datasets.
% \shijie{This sentence seems too complex, try to rephrase:}
Motivated by the success of contrastive vision-language models and the curiosity of the language effect mentioned above, we directly test CLIP on LTR datasets under its zero-shot setting.
% integrate text embeddings of categories into visual features following zero-shot inference of CLIP on LTR datasets. 
Surprisingly, the results are balanced on many-shots ($59.4\%$), medium-shots ($57.5\%$), and low-shots ($57.6\%$) subsets of ImageNet-LT~\cite{liu2019large} and the overall performance ($58.2\%$) is comparable to the state-of-the-art~\cite{cui2021parametric}. From which we see the great potential of the multimodality solution for LTR. 
% \jiasen{We need to show clip zero-shot number.}
To further improve the performance while keep the capability of dealing with data imbalance, an intuitive way is to finetune the vision-language models on LTR datasets. However, we find it only brings a slight gain. Therefore, the core task of our work becomes \textit{how to design an effective recipe for training vision-language models under the circumstances of long-tailed distribution.}

%Motivated by the success of contrastive vision-language models, we directly test CLIP with ResNet-50 visual backbone on LTR datasets and achieve slightly lower average performance than the state-of-the-art approach PaCo~\cite{cui2021parametric} under zero-shot setting. By diving deeper into different criteria, we discover that CLIP can achieve consistent performances
% \jiasen{consistent performance?}
%on many-shots, medium-shots and low-shots setups. In contrast, PaCo obtains good results 
% \jiasen{state of the art results or good results?}
%on many-shots yet significant worse performance 
% \jiasen{significant worse performance?}
%on few-shots due to imbalanced class distribution.

% The above observation reveals the potential of contrastive vision-language models on LTR \jiasen{Maybe it is better to put this sentence to the last paragraph and explain the potential a little bit more}.
%In view of above observations, 
Specifically, in this paper, we design a simple framework based on contrastive vision-language models for LTR. The training procedure of the framework is broken into two phases from the perspective of distribution skewness: A) utilizing abundant annotations from LTR datasets; B) tackling few-shot learning of tail classes on balanced data built with re-sampling strategies. In Phase A, we continue pretraining CLIP backbone on a specific LTR dataset through contrastive learning. It enables our framework to fully exploit available training examples and update visual-language representations on a new domain.
% After Phase A, head classes typically achieve good enough classification performance, whereas tail classes are still far from perfect. 
To further facilitate the few-shot learning of tail classes, during Phase B, we freeze the CLIP backbone and employ an auxiliary linear adapter for finetuning on re-balanced training samples. The adapter dynamically combines fixed Phase-A and finetuned Phase-B features via a residual connection to refine the visual representations of tail classes. Compared with finetuning the whole CLIP backbone directly, the linear adapter reduces the number of learnable parameters and thus prevents the potential overfitting of few-shot setups.
% To further facilitate the few-shot learning of tail classes, during Phase B, we first adopt re-sampling strategies to build balanced training samples. 
% We then freeze the CLIP backbone and employ an additional adapter for fine-tuning on the balanced data. 
% The adapter is a lightweight linear layer over the visual branch of CLIP. 
% It combines fixed Phase-A and fine-tuned Phase-B visual features via residual connections. Compared with fine-tuning CLIP backbone directly, linear adapter reduces the number of learnable parameters and thus prevents potential overfitting of few-shot setups. 
% After Phase B, tail classes enjoy a performance boost while head classes slightly increase or maintain their original classification accuracy. 
% \jiasen{This paragraph contains too much technique details. We need to simply the paragraph and point out more high level intuition between these two phrase}
According to Figure~\ref{fig:intro}, our framework clearly achieves better performances than state-of-the-art LTR approaches.~The improvements are especially significant for few-shot and medium-shot classes, demonstrating our approach's great capability of handling class imbalance. Since our framework solves the data imbalance via a linear adapter, we name it as \approach{} (\textbf{BAL}anced \textbf{L}inear \textbf{AD}apter), which implies the harmony of head and tail classes.
Our contributions are three folds:
\begin{itemize}[leftmargin=*,noitemsep] 
    \item We point out the shortcomings of training with fixed class labels and propose to leverage language modality via contrastive vision-language backbone to facilitate long-tailed recognition.
    \item We develop the \approach{} framework consisting of two phases to handle head and tail classes successively. Specifically, we keep training the visual and language branches of the pretrained vision-language model simultaneously at the first stage. Then we adopt a linear adapter to tackle tail classes with vision-language parameters frozen.
    %we adopt a simple linear adapter to tackle the few-shot learning of tail classes.
    \item We conduct extensive experiments to demonstrate the effectiveness of \approach{}. Our simple baseline achieves the new state-of-the-art performances on all benchmarks, outperforming the old paradigm by $16.5$ points maximally on ImageNet-LT.
    %We include our code in supplementary materials and will release it soon.
    % The code will be released at {\url{https://github.com/xxx/xxx}}.
\end{itemize}

\section{Related Work}
\label{sec:related}
\noindent\textbf{Contrastive Vision-Language Model.~}
Contrastive representation learning has been widely adopted to fulfill self-supervised pretraining in various AI domains\cite{chen2020simple,he2020momentum,caron2020unsupervised,caron2021emerging,oord2018representation,gao2021simcse}. 
%It aims to learn a representation space where semantically similar examples stay close to each other while disparate ones stand away.
Recently, the intersection of vision and language~\cite{antol2015vqa,nguyen2018improved,yu2019deep,gao2019dynamic,kim2018bilinear,chen2019uniter,shi2020contrastive} also experienced a revolution sparked by contrastive representation learning. Contrastive vision-language models like CLIP~\cite{radford2021learning} and ALIGN~\cite{jia2021scaling} demonstrate promising zero-shot performances on various visual search and recognition tasks. Learning directly from natural language supervisions that contain rich visual concepts, they are very flexible and robust to distribution variations across different domains. The success of CLIP and ALIGN has enlightened many downstream vision-language tasks. For instance, DeCLIP~\cite{li2021supervision} proposes to utilize self-, multi-view, and nearest-neighbor supervisions among the image-text pairs for data efficient pretraining of CLIP. On visual classification tasks, 
%CoOp~\cite{zhou2021coop} adopts soft learnable prompts and conducts prompt tuning to improve contrastive vision-language models in terms of few-shot image classification. 
CLIP-Adapter~\cite{gao2021clip} argues that fine-tuning contrastive vision-language models with linear adapters is a better alternative to prompt tuning. For video related tasks, VideoCLIP~\cite{xu2021videoclip} performs contrastive pretraining with video-text pairs for zero-shot video-text understanding. ActionCLIP~\cite{wang2021actionclip} presents a new ``pretrain, prompt and fine-tune'' paradigm leveraging pretrained vision-language models for zero-shot/few-shot action recognition. CLIP-It~\cite{narasimhan2021clip} designs a language-guided multimodal transformer based on CLIP to address query-focused video summarization. Moreover, CLIPort~\cite{shridhar2021cliport} combines CLIP with Transporter~\cite{zeng2020transporter} to endow a robot with the ability of semantic understanding and spatial perception. In this paper, we demonstrate that contrastive vision-language models can also facilitate visual recognition under long-tailed class distribution setups if properly trained.

\noindent\textbf{Long-Tailed Recognition.~}
% \shijie{Two aspects -- 1) datasets: image classification (iNaturalist) and instance segmentation (LVIS); 2) different long-tailed recognition methods.}
Long-tailed recognition~\cite{zhang2021deep} is a practical and challenging problem in vision domain. General visual models will suffer from severe performance degradation under such imbalanced class distributions.
A great number of approaches~\cite{huang2016learning,ouyang2016factors,zhang2017range,dong2017class,yin2019feature,menon2021longtail,cui2021parametric,wang2021longtailed,Samuel_2021_ICCV,cui2021reslt} have been proposed to address LTR from different perspectives. An intuitive solution is to directly re-balance the number of training samples across all classes~\cite{Kang2020Decoupling,zhou2020bbn}. However, naively adjusting the skewness of training samples may lead to the overfitting of tail classes. Better alternatives include loss re-weighting~\cite{lin2017focal,kang2019few,hong2021disentangling} and logit adjustment~\cite{menon2021longtail,zhang2021distribution} based on label frequencies. Though efficacious for long-tailed distribution, above methods all sacrifice the performance of head classes at 
varying levels. To address the limitations, researchers turn to explore new network architectures and training paradigms. Typically, long-tail recognition models contain two 
key components -- feature extractor and classifier. For each component, there are corresponding approaches by either designing better classifier~\cite{liu2020deep,tang2020long,wu2020solving} or learning reliable representations~\cite{liu2019large,zhu2020inflated}.
% \shijie{TBD: network architectures}
In terms of new training frameworks, existing efforts seek to divide a one-stage training paradigm into two stages. For example, decoupled training approaches~\cite{Kang2020Decoupling,kang2021exploring} conduct representation learning and classifier training in a separate manner.
Furthermore, ensemble learning schemes~\cite{zhou2020bbn,zhang2021test} first learn multiple experts with different data sub-groups and then merge their complementary knowledge to handle LTR.
% Transfer learning methods generally attempt to transfer learned knowledge from head classes, a source domain, or a teacher model to tail classes, a specific target domain, or a student model, respectively.
% \shijie{two-stage: ensemble learning (experts + ensemble), transfer learning (knowledge + transfer to student, tail classes, specific domain)}
% new network architectures: representation learning (feature extractor), classifier design
% new training frameworks: decoupled training, ensemble learning, transfer learning
% \shijie{Include different two-stage methods.}
% Two-stage methods have been widely adopted in data imbalanced and domain adaptation situations.
% Lu~\etal~\cite{lu2021simpler} designs a few-shot learning approach to address long-tailed semantic segmentation.
% 具体哪里不同? -> 只有representation learning, 分成两阶段: 充分利用数据, 鼓励继续学习细粒度特征
In contrast, our \approach{} first utilizes abundant long-tailed data to refine visual-language representations on a new target domain. Then we apply a lightweight linear adapter to encourage fine-grained representation learning from balanced samples. The two phases successively handle the learning of head and tail classes and ensure a better balanced performance across all classes.
% learning representation equals to refining both visual feature and classifier, which is very different from decoupled training approaches.

\section{Our Method}
\label{method}
In this section, we first briefly revisit how contrastive vision-language models leverage contrastive objectives to achieve efficient and scalable multimodal representation learning. Moreover, we formally present \approach{} framework and discuss the advantages of the proposed two-stage representation learning for long-tailed class distributions.

\subsection{Contrastive Vision-Language Model}
% visual/text encoders -> L2 norm
Contrastive vision-language models such as CLIP~\cite{radford2021learning} and ALIGN~\cite{jia2021scaling} typically follow a dual-encoder architecture with a language encoder $\mathcal{L}_{\mathrm{enc}}$ and a visual encoder $\mathcal{V}_{\mathrm{enc}}$.
Given an input image $\bm{I}$, $\mathcal{V}_{\mathrm{enc}}$ is adopted to extract the visual feature for $\bm{I}$: $\bm{f}_v = \mathcal{V}_{\mathrm{enc}}(\bm{I}) \in \mathbb{R}^{d_v}$. Likewise, $\mathcal{L}_{\mathrm{enc}}$ is applied to encode an input text sequence $\bm{T}$ into its corresponding text feature: $\bm{f}_l = \mathcal{L}_{\mathrm{enc}}(\bm{T}) \in \mathbb{R}^{d_l}$. After extracting the feature for each modality, two transformation matrices $\mathbf{W}_{v} \in \mathbb{R}^{d_v\times d}$ and $\mathbf{W}_{l} \in \mathbb{R}^{d_l\times d}$ are employed to project the original visual and text features into a shared embedding space: 
\begin{equation}
\label{eq:joint}
\mathbf{v} = \frac{\mathbf{W}_{v}^{\top}\bm{f}_v}{\|\mathbf{W}_{v}^{\top}\bm{f}_v\|}, \ \ \  \mathbf{u} = \frac{\mathbf{W}_{l}^{\top}\bm{f}_l}{\|\mathbf{W}_{l}^{\top}\bm{f}_l\|},
\end{equation}
where $\mathbf{v}$ and $\mathbf{u}$ are both $d$-dimension normalized vectors in the joint multimodal space.
% training objectives
During pretraining, contrastive vision-language models learn to align image-text pairs inside a batch. The overall training objective consists of matching losses from two different directions, \emph{i.e.}, $\mathscr{L}_{v\rightarrow l}$ for text retrieval and $\mathscr{L}_{l\rightarrow v}$ for image retrieval. They both maximize the scores of matched pairs while minimize that of unmatched ones:
\begin{align}
\mathscr{L}_{v\rightarrow l} & = -\frac{1}{N} \sum_{i}^{N} \log \frac{\exp \left(\mathbf{v}_{i}^{\top} \mathbf{u}_{i} / \tau\right)}{\sum_{j=1}^{N} \exp \left(\mathbf{v}_{i}^{\top} \mathbf{u}_{j} / \tau\right)}, \label{eq:v2l}\\
\mathscr{L}_{l\rightarrow v} & =-\frac{1}{N} \sum_{i}^{N} \log \frac{\exp \left(\mathbf{u}_{i}^{\top} \mathbf{v}_{i} / \tau\right)}{\sum_{j=1}^{N} \exp \left(\mathbf{u}_{i}^{\top} \mathbf{v}_{j} / \tau\right)}, \label{eq:l2v}
\end{align}
where $\tau$ denotes the temperature hyperparameter and $N$ represents the number of image-text pairs in the batch.

% how to conduct classification
Trained with large-scale image-text pairs under the open-vocabulary settings, contrastive vision-language models achieve powerful multimodal representations and naturally possess the capability of zero-shot visual recognition. A collection of text descriptions following templates like ''a photo of a \{\texttt{CLASS}\}'' is created for candidate classes in target datasets to perform zero-shot prediction. If we represent the normalized test image feature as $\mathbf{v}$ and all normalized text description features as $\{\mathbf{u}_1, \cdots, \mathbf{u}_K\}$, we can thus compute the class probability of the test image as below:
\begin{equation}
p_{i}=\frac{\exp \left(\mathbf{v}^{\top} \mathbf{u}_i\right) / \tau}{\sum_{j=1}^{K} \exp \left(\mathbf{v}^{\top} \mathbf{u}_j\right) / \tau},
\end{equation}
where $p_i$ represents the probability for class $i$, and $K$ stands for the total number of candidate classes. Finally, the text label with the highest probability is selected as the prediction.
% $\mathop{\arg\max}_{i} p_{i}$ 

\subsection{Balanced Linear Adapter}
% why two stages?
As stated in Section~\ref{sec:intro}, contrastive vision-language models obtain balanced performance for head and tail classes, whereas traditional approaches like PaCo~\cite{cui2021parametric} suffer from lower performance of tail classes owing to the deficiency of training samples. Inspired by the zero-shot ability of contrastive vision-language models, we choose CLIP as our backbone for long-tailed recognition. The observation in Section~\ref{subsec:adapter} also encourages us to decouple the training of long-tailed data into two phases. To be specific, the first phase (Phase A) fully utilizes available training data and ensures the performance for classes with abundant examples, then the second phase (Phase B) focuses on improving the few-shot learning of tail classes. Note that LWS~\cite{Kang2020Decoupling} also adopts a decoupled training framework. However, LWS decouples the training of representation and classifier into two stages. In contrast, our two phases are for long-tailed and balanced training samples respectively and both phases conduct representation refinement with contrastive loss.

% advantage of don't stop pretraining 
% Phase A: fully utilizes available training data.
\noindent\textbf{Phase A.~} Recently, Gururangan \etal~\cite{gururangan2020don} shows that keeping domain-adaptive and task-adaptive model pretraining can largely improve the performances on target NLP tasks. Similarly, for our Phase A, we find that continuing the pretraining of contrastive vision-language backbone on long-tailed target dataset also benefits the learning of classes with abundant examples. In this way, Phase A can make full use of available training data regardless of its skewness. Since we focus on classifying input images into text labels, the pretraining of Phase A directly follows the loss defined in Equation (\ref{eq:v2l}). As shown in Figure~\ref{fig:1}, Phase A updates the representations of both text and image encoders on a new domain. After Phase A, head classes typically achieve good performance while tail classes still require another stage of balanced training.

% Phase B: encourage representation refinement on tail classes (usually rare or fine-grained).
\noindent\textbf{Phase B.~} Tail classes are short of training examples and under the few-shot settings. Directly training the whole vision-language backbone may easily overfit to them and lead to performance degradation. Inspired by parameter-efficient adapter modules~\cite{houlsby2019parameter}, we freeze the vision-language backbone obtained from Phase A  and utilize an additional linear adapter layer to help our model refine its visual-language representation on those infrequent classes. 
% introduce linear adapter -> how to conduct fine-tuning (text prompt)
As shown in Figure~\ref{fig:1}, the text features would remain the same as Phase A. The only difference lies in the image features. If we assume the original image feature to be $\bm{f}$, the weight matrix and bias of the linear adapter as $\mathbf{W} \in \mathbb{R}^{d \times d}$ and $\mathbf{b} \in \mathbb{R}^{d}$, then we can represent the refined image feature $\bm{f}^{\star}$ as
\begin{equation}
\bm{f}^{\star}=\lambda\cdot\operatorname{ReLU}\left(\mathbf{W}^{\top}\bm{f} + \mathbf{b}\right) + (1-\lambda)\cdot \bm{f},
\end{equation}
where $\lambda$ indicates the residual factor to dynamically combine Phase-B fine-tuned image features with the original image features of Phase A.

To avoid the Phase-B training from biasing towards head classes, we also adopt class-balanced sampling strategy~\cite{Kang2020Decoupling} to construct a balanced group of training samples. Suppose there are $K$ classes that constitute a total of $N$ training samples in the target dataset. We can represent the number of training samples for class $j$ as $n_j$ and thus have $N = \sum_{j=1}^{K}n_j$. If we assume these classes are already sorted in a decreasing order, then a long-tailed distribution implies $n_i \ge n_j$ if $i < j$ and $n_1 \gg n_K$. For class-balanced sampling, we define the probability of sampling each data point from class $j$ to be $q_{j}=\frac{1}{K}$. In other words, to construct a balanced group of training samples, we will first uniformly choose a class out of the $K$ candidates and then sample one data point from the selected class. Finally, we perform Phase B finetuning with $\mathscr{L}_{v\rightarrow l}$ on the balanced training data.

\section{Experiments}
\subsection{Experiment Setup}
\noindent\textbf{Datasets.~}
We conduct our experiments on three long-tailed benchmark datasets, namely ImageNet-LT~\cite{liu2019large}, Places-LT~\cite{liu2019large}, and CIFAR100-LT~\cite{cao2019learning}. ImageNet-LT and Places-LT were first introduced in~\cite{liu2019large} for long-tailed recognition research. ImageNet-LT is a long-tailed dataset with 1,000 categories sampled from the original ImageNet~\cite{deng2009imagenet} following the Pareto distribution with 
a power value of $\alpha = 6$. There are 115.8K images in the training split, with maximally 1,280 images per class and minimally 5 images per class. The testing split maintains the same as the original ImageNet~\cite{deng2009imagenet}, where samples per class are balanced.
Places-LT is a long-tailed version of the original Places2 Database~\cite{zhou2017places}. The training split of  Places-LT contains with 184.5K images from 365 categories, with 4,980 images maximally per class and minimally 5 images per class. For the testing split, the images of each class is also balanced with 100 images per class. CIFAR100-LT~\cite{cao2019learning} are created by long-tailed imbalance technique~\cite{cui2019class} which reduces training examples per class based on an exponential decay function. In this paper, we directly use the version from~\cite{wang2021longtailed} with an imbalance ratio $\rho$ of $100$. The training split contains 50K images from 100 categories, while the testing split has a uniform 100 images for each class.

\noindent\textbf{Implementation Details.~}
We use CLIP as the contrastive vision-language backbone in all experiments. For the visual branch of CLIP, we vary among ResNet-50, ResNet-100, ViT-B/16, and ResNet-50$\times$16, which is 16$\times$ computation cost of ResNet-50 following the style of EfficientNet as introduced in~\cite{radford2021learning}. The ResNet-50 is leveraged for all ablation studies by default unless specified. We use SGD as the optimizer for all experiments with a momentum of $0.9$. The batch size is set to $512$. We adopt cosine learning rate schedule to decay learning rates. The initial learning rate of CLIP finetuning is set to $1\times 10^{-5}$
% 0.00001
for both the visual and language encoders, while the learning rate of linear adapter is set to $0.2$ at the start. For data pre-processing, images are resized to $224 \times 224$ unless the ResNet-50$\times$16 visual backbone, which utilizes an image size of $384 \times 384$. Crop and random horizontal flip are also adopted to augment the original images for robustness considerations. For \approach{}, we train Phase A for 50 epochs and and Phase B for 10 epochs by default unless specified. For the hyperparameters, we set the residual factor $\lambda$ to $0.2$ and the temperature $\tau$ to $1.0$. The feature dimensions of ResNet-50, ResNet-101, ResNet-50$\times$16 and ViT-B/16 are $1024, 512, 768, 512$ respectively.
%d_l和d_v一样，介绍一下总的即可
% \shijie{Add dimension of $d$, $d_v$, $d_l$?}

\noindent\textbf{Evaluation Metrics.~}
We evaluate the models for long-tailed recognition on the balanced test splits and report the commonly used top-1 classification accuracy of \textit{all} classes. Following~\cite{Kang2020Decoupling}, we divide these classes into three subsets -- \textit{many-shot}, \textit{medium-shot}, and \textit{few-shot} categories.~Specifically, \textit{many-shot}, \textit{medium-shot}, and \textit{few-shot} are decided according to the amount of instances in each category, namely more than 100 images, 20-100 images, and less than 20 images, respectively.

\begin{table}[t!]
    \centering
    \resizebox{\columnwidth}{!}{
    \begin{tabular}{clllllll}
    \toprule
         \multicolumn{1}{c|}{Method}  &\multicolumn{1}{c|}{Backbone} &\multicolumn{1}{c|}{\#Epochs} &\multicolumn{1}{c}{Many} &\multicolumn{1}{c}{Medium} &\multicolumn{1}{c}{Few} &\multicolumn{1}{c}{All}\\ \midrule
         \multirow{6}*{$\tau$-normalized~\cite{Kang2020Decoupling}}  &\multicolumn{1}{|c|}{ResNet-50} &\multicolumn{1}{c|}{90} &\multicolumn{1}{c}{56.6} &\multicolumn{1}{c}{44.2} &\multicolumn{1}{c}{27.4} &\multicolumn{1}{c}{46.7} \\
         ~&\multicolumn{1}{|c|}{ResNet-101} &\multicolumn{1}{c|}{90} &\multicolumn{1}{c}{59.4} &\multicolumn{1}{c}{47.0} &\multicolumn{1}{c}{30.6} &\multicolumn{1}{c}{49.6} \\
         ~&\multicolumn{1}{|c|}{ResNet-152} &\multicolumn{1}{c|}{90} &\multicolumn{1}{c}{59.6} &\multicolumn{1}{c}{47.5} &\multicolumn{1}{c}{32.2} &\multicolumn{1}{c}{50.1} \\
           ~ &\multicolumn{1}{|c|}{ResNeXt-50} &\multicolumn{1}{c|}{90} &\multicolumn{1}{c}{59.1} &\multicolumn{1}{c}{46.9} &\multicolumn{1}{c}{30.7} &\multicolumn{1}{c}{49.4} \\ 
           ~ &\multicolumn{1}{|c|}{ResNeXt-101} &\multicolumn{1}{c|}{90} &\multicolumn{1}{c}{59.1} &\multicolumn{1}{c}{47.0} &\multicolumn{1}{c}{31.7} &\multicolumn{1}{c}{49.6} \\ 
           ~ &\multicolumn{1}{|c|}{ResNeXt-152} &\multicolumn{1}{c|}{90} &\multicolumn{1}{c}{62.2} &\multicolumn{1}{c}{50.1} &\multicolumn{1}{c}{35.8} &\multicolumn{1}{c}{52.8} \\
    \midrule
         \multirow{6}*{LWS~\cite{Kang2020Decoupling}}  &\multicolumn{1}{|c|}{ResNet-50} &\multicolumn{1}{c|}{90} &\multicolumn{1}{c}{57.1} &\multicolumn{1}{c}{45.2} &\multicolumn{1}{c}{29.3} &\multicolumn{1}{c}{47.7} \\
         ~&\multicolumn{1}{|c|}{ResNet-101} &\multicolumn{1}{c|}{90} &\multicolumn{1}{c}{60.1} &\multicolumn{1}{c}{47.6} &\multicolumn{1}{c}{31.2} &\multicolumn{1}{c}{50.2} \\
         ~&\multicolumn{1}{|c|}{ResNet-152} &\multicolumn{1}{c|}{90} &\multicolumn{1}{c}{60.6} &\multicolumn{1}{c}{47.8} &\multicolumn{1}{c}{31.4} &\multicolumn{1}{c}{50.5} \\
           ~ &\multicolumn{1}{|c|}{ResNeXt-50} &\multicolumn{1}{c|}{90} &\multicolumn{1}{c}{60.2} &\multicolumn{1}{c}{47.2} &\multicolumn{1}{c}{30.3} &\multicolumn{1}{c}{49.9} \\ 
           ~ &\multicolumn{1}{|c|}{ResNeXt-101} &\multicolumn{1}{c|}{90} &\multicolumn{1}{c}{60.5} &\multicolumn{1}{c}{47.2} &\multicolumn{1}{c}{31.2} &\multicolumn{1}{c}{50.1} \\ 
           ~ &\multicolumn{1}{|c|}{ResNeXt-152} &\multicolumn{1}{c|}{90} &\multicolumn{1}{c}{63.5} &\multicolumn{1}{c}{50.4} &\multicolumn{1}{c}{34.2} &\multicolumn{1}{c}{53.3} \\
    \midrule
        \multirow{2}*{ResLT~\cite{cui2021reslt}}  &\multicolumn{1}{|c|}{ResNeXt-50} &\multicolumn{1}{c|}{180} &\multicolumn{1}{c}{63.0} &\multicolumn{1}{c}{50.5} &\multicolumn{1}{c}{35.5} &\multicolumn{1}{c}{52.9} \\
           ~ &\multicolumn{1}{|c|}{ResNeXt-101} &\multicolumn{1}{c|}{180} &\multicolumn{1}{c}{63.3} &\multicolumn{1}{c}{53.3} &\multicolumn{1}{c}{40.3} &\multicolumn{1}{c}{55.1} \\ 
    \midrule
         \multirow{3}*{Balanced Softmax~\cite{ren2020balanced}}  &\multicolumn{1}{|c|}{ResNet-50} &\multicolumn{1}{c|}{400} &\multicolumn{1}{c}{66.7} &\multicolumn{1}{c}{52.9} &\multicolumn{1}{c}{33.0} &\multicolumn{1}{c}{55.0} \\
           ~ &\multicolumn{1}{|c|}{ResNeXt-50} &\multicolumn{1}{c|}{400} &\multicolumn{1}{c}{67.7} &\multicolumn{1}{c}{53.8} &\multicolumn{1}{c}{34.2} &\multicolumn{1}{c}{56.2} \\ 
           ~ &\multicolumn{1}{|c|}{ResNeXt-101} &\multicolumn{1}{c|}{400} &\multicolumn{1}{c}{69.2} &\multicolumn{1}{c}{55.8} &\multicolumn{1}{c}{36.3} &\multicolumn{1}{c}{58.0} \\ 
    \midrule
         \multirow{2}*{RIDE$\dagger$~\cite{wang2020long}}  &\multicolumn{1}{|c|}{ResNet-50} &\multicolumn{1}{c|}{100} &\multicolumn{1}{c}{66.2} &\multicolumn{1}{c}{52.3} &\multicolumn{1}{c}{36.5} &\multicolumn{1}{c}{55.4} \\
           ~ &\multicolumn{1}{|c|}{ResNeXt-50} &\multicolumn{1}{c|}{100} &\multicolumn{1}{c}{68.2} &\multicolumn{1}{c}{53.8} &\multicolumn{1}{c}{36.0} &\multicolumn{1}{c}{56.8} \\ 
    \midrule
         \multirow{3}*{PaCo$\ddagger$~\cite{cui2021parametric}}  &\multicolumn{1}{|c|}{ResNet-50} &\multicolumn{1}{c|}{400} &\multicolumn{1}{c}{65.0} &\multicolumn{1}{c}{55.7} &\multicolumn{1}{c}{38.2} &\multicolumn{1}{c}{57.0} \\
           ~ &\multicolumn{1}{|c|}{ResNeXt-50} &\multicolumn{1}{c|}{400} &\multicolumn{1}{c}{67.5} &\multicolumn{1}{c}{56.9} &\multicolumn{1}{c}{36.7} &\multicolumn{1}{c}{58.2} \\ 
           ~ &\multicolumn{1}{|c|}{ResNeXt-101} &\multicolumn{1}{c|}{400} &\multicolumn{1}{c}{68.2} &\multicolumn{1}{c}{58.7} &\multicolumn{1}{c}{41.0} &\multicolumn{1}{c}{60.0} \\ 
    \midrule
         \multirow{4}*{\approach{}}  &\multicolumn{1}{|c|}{ResNet-50} &\multicolumn{1}{c|}{50+10} &\multicolumn{1}{c}{71.0} &\multicolumn{1}{c}{66.3} &\multicolumn{1}{c}{59.5} &\multicolumn{1}{c}{67.2 \textcolor{red}{(+7.2)}} \\
           ~ &\multicolumn{1}{|c|}{ResNet-101} &\multicolumn{1}{c|}{50+10} &\multicolumn{1}{c}{74.7} &\multicolumn{1}{c}{69.1} &\multicolumn{1}{c}{63.3} &\multicolumn{1}{c}{70.5 \textcolor{red}{(+10.5)}} \\ 
           ~ &\multicolumn{1}{|c|}{ViT-B/16} &\multicolumn{1}{c|}{50+10} &\multicolumn{1}{c}{79.1} &\multicolumn{1}{c}{74.5} &\multicolumn{1}{c}{\textbf{69.8}} &\multicolumn{1}{c}{75.7 \textcolor{red}{(+15.7)}} \\
           ~ &\multicolumn{1}{|c|}{ResNet-50$\times$16} &\multicolumn{1}{c|}{50+10} &\multicolumn{1}{c}{\textbf{81.1}} &\multicolumn{1}{c}{\textbf{75.6}} &\multicolumn{1}{c}{67.0} &\multicolumn{1}{c}{\textbf{76.5} \textcolor{red}{\bf (+16.5)}} \\ 
    \bottomrule
    \end{tabular}
    }
    \caption{Long-tailed recognition accuracy on ImageNet-LT for different methods and backbones. The red colored numbers represent the improvement of overall accuracy compared with the state-of-the-art performance (PaCo with $60.0$\% overall accuracy). $\dagger$: RIDE uses 4 experts. $\ddagger$: the state-of-the-art model.}
    \vspace{-8pt}
    \label{tab:imagenet}
\end{table}

\subsection{Performance Comparison}
In this section, we compare the performance of \approach{} with long-tailed recognition approaches that report state-of-the-art results on three benchmark datasets, \emph{i.e.}, ImageNet-LT, Places-LT, and CIFAR100-LT.

\noindent\textbf{ImageNet-LT.~}
Table~\ref{tab:imagenet} shows the long-tailed recognition results on ImageNet-LT. We present \approach{} variants with ResNet-50, ResNet-101, ResNet-50$\times$16, and ViT-B/16 as the visual backbone. From the table, we observe that with only $50+10$ epochs ($50$ epochs for Phase A and $10$ epochs for Phase B), our smallest \approach{} variant with ResNet-50 visual backbone can surpass the largest model of the state-of-the-art PaCo~\cite{cui2021parametric} using ResNeXt-101 by $+7.2\%$. When gradually increasing the size of visual backbone, we find the performance of \approach{} also enjoys an improvement. It is worth noting that \approach{} with ResNet-50$\times$16 achieves an accuracy of $76.5\%$, which outperforms other state-of-the-art models with a large margin.
% We further experiment with adding the compute of backbone to raise the performance of \approach{}, as is shown in Table~\ref{tab:imagenet},  our maximal model using 16x the compute of ResNet-50 following Efficient-Net style can achieve up to $76.5\%$, outperforming all other state-of-the-art models with a great margin. 

\begin{table}[t!]
    \centering
    \resizebox{\columnwidth}{!}{
    \begin{tabular}{clllllll}
    \toprule
         \multicolumn{1}{c|}{Method}  &\multicolumn{1}{c|}{Backbone} &\multicolumn{1}{c|}{\#Pretrain} &\multicolumn{1}{c}{Many} &\multicolumn{1}{c}{Medium} &\multicolumn{1}{c}{Few} &\multicolumn{1}{c}{All}\\ \midrule
         \multirow{1}*{OLTR~\cite{liu2019large}}  &\multicolumn{1}{|c|}{ResNet-152} &\multicolumn{1}{c|}{Y} &\multicolumn{1}{c}{44.7} &\multicolumn{1}{c}{37.0} &\multicolumn{1}{c}{25.3} &\multicolumn{1}{c}{35.9} \\  
         
         \multirow{1}*{cRT~\cite{Kang2020Decoupling}}  &\multicolumn{1}{|c|}{ResNet-152} &\multicolumn{1}{c|}{Y} &\multicolumn{1}{c}{42} &\multicolumn{1}{c}{37.6} &\multicolumn{1}{c}{24.9} &\multicolumn{1}{c}{36.7} \\
         
         \multirow{1}*{$\tau$-normalized~\cite{Kang2020Decoupling}}  &\multicolumn{1}{|c|}{ResNet-152} &\multicolumn{1}{c|}{Y} &\multicolumn{1}{c}{37.8} &\multicolumn{1}{c}{40.7} &\multicolumn{1}{c}{31.8} &\multicolumn{1}{c}{37.9} \\

         \multirow{1}*{LWS~\cite{Kang2020Decoupling}}  &\multicolumn{1}{|c|}{ResNet-152} &\multicolumn{1}{c|}{Y} &\multicolumn{1}{c}{40.6} &\multicolumn{1}{c}{39.1} &\multicolumn{1}{c}{28.6} &\multicolumn{1}{c}{37.6} \\
         
         \multirow{1}*{Balanced Softmax~\cite{ren2020balanced}}  &\multicolumn{1}{|c|}{ResNet-152} &\multicolumn{1}{c|}{Y} &\multicolumn{1}{c}{42.0} &\multicolumn{1}{c}{39.3} &\multicolumn{1}{c}{30.5} &\multicolumn{1}{c}{38.6} \\

        \multirow{1}*{ResLT~\cite{cui2021reslt}}  &\multicolumn{1}{|c|}{ResNet-152} &\multicolumn{1}{c|}{Y} &\multicolumn{1}{c}{39.8} &\multicolumn{1}{c}{43.6} &\multicolumn{1}{c}{31.4} &\multicolumn{1}{c}{39.8} \\

        % \multirow{1}*{RIDE$\dagger$~\cite{wang2020long}}  &\multicolumn{1}{|c|}{ResNet-152} &\multicolumn{1}{c|}{Y} &\multicolumn{1}{c}{66.2} &\multicolumn{1}{c}{52.3} &\multicolumn{1}{c}{36.5} &\multicolumn{1}{c}{55.4} \\
        \midrule
         \multirow{1}*{PaCo~\cite{cui2021parametric}}  &\multicolumn{1}{|c|}{ResNet-152} &\multicolumn{1}{c|}{Y} &\multicolumn{1}{c}{37.5} &\multicolumn{1}{c}{47.2} &\multicolumn{1}{c}{33.9} &\multicolumn{1}{c}{41.2} \\
         \multirow{1}*{PaCo$\dagger$~\cite{cui2021parametric}} &\multicolumn{1}{|c|}{ResNet-152} &\multicolumn{1}{c|}{Y} &\multicolumn{1}{c}{36.1} &\multicolumn{1}{c}{47.9} &\multicolumn{1}{c}{35.3} &\multicolumn{1}{c}{41.2} \\
         
    \midrule
         \multirow{4}*{\approach{}}  &\multicolumn{1}{|c|}{ResNet-50} &\multicolumn{1}{c|}{N} &\multicolumn{1}{c}{46.7} &\multicolumn{1}{c}{48.0} &\multicolumn{1}{c}{42.7} &\multicolumn{1}{c}{46.5 \textcolor{red}{(+5.3)}} \\
           ~ &\multicolumn{1}{|c|}{ResNet-101} &\multicolumn{1}{c|}{N} &\multicolumn{1}{c}{48.0} &\multicolumn{1}{c}{48.6} &\multicolumn{1}{c}{46.0} &\multicolumn{1}{c}{47.9 \textcolor{red}{(+6.7)}} \\ 
           ~ &\multicolumn{1}{|c|}{ViT-B/16} &\multicolumn{1}{c|}{N} &\multicolumn{1}{c}{49.3} &\multicolumn{1}{c}{50.2} &\multicolumn{1}{c}{\textbf{48.4}} &\multicolumn{1}{c}{\textbf{49.5} \textcolor{red}{\bf (+8.3)}} \\
           ~ &\multicolumn{1}{|c|}{ResNet-50$\times$16} &\multicolumn{1}{c|}{N} &\multicolumn{1}{c}{\textbf{49.4}} &\multicolumn{1}{c}{\textbf{50.5}} &\multicolumn{1}{c}{46.6} &\multicolumn{1}{c}{49.3 \textcolor{red}{(+8.1)}} \\ 
    \bottomrule
    \end{tabular}
    }
    \caption{Long-tailed recognition accuracy on Places-LT for different methods. The red colored numbers represent improvement of overall accuracy compared with the state-of-the-art performance (PaCo with $41.2$\% overall accuracy). \#Pretrain: whether pretrain visual backbone on full ImageNet-2012 or not. $\dagger$: PaCo variant trained with RandAugment~\cite{cubuk2020randaugment}.}
    \vspace{-10pt}
    \label{tab:places}
\end{table}

\noindent\textbf{Places-LT.~}
We further evaluate \approach{} on Places-LT dataset and report the results in Table~\ref{tab:places}.
It is a commonly used scheme of previous approaches to pretrain their models on ImageNet-2012 full dataset first to enrich the visual representation before finetuning on Places-LT.
% While previous approaches require ImageNet-2012 pretraining as a prerequisite, 
However, \approach{} can directly perform training on Places-LT thanks to the additional language representation of contrastive vision-language models. 
%\approach{} is capable of prompting rich features obtained from contrastive learning via text embeddings of categories in Places-LT.
% It is a commonly used scheme to pretrain the visual backbone on ImageNet-2012 full dataset to enrich the visual representation before finetuning on Places-LT. 
% Thanks to the powerful visual-language representation of contrastive vision-language models, \approach{} does not need to be pretrained on other datasets except Places-LT. 
% \approach{} is capable of prompting rich features obtained from contrastive learning via text embeddings of categories in Places-LT. 
% However, Vision-Language model of \approach{} has no necessity to be pretrained on other datasets except Places-LT, as \approach{} is capable of prompting rich features obtained from contrastive learning via text embeddings of categories in Places-LT. 
As shown in Table~\ref{tab:places}, \approach{} with ResNet-50 visual backbone achieves $46.5\%$ accuracy for \textit{all} categories, which beats the state-of-the-art model PaCo with ResNet-152 by $+5.3\%$. This shows \approach{} can not only achieve better performance with smaller visual backbone but also save a great amount of training time by skipping the ImageNet pretraining.
% The best \approach{} variant on Places-LT is with ViT-B/16 visual backbone and achieves the highest $48.4\%$ accuracy for \textit{few-shot} categories and $49.5\%$ accuracy for \textit{all} categories.

\begin{table}[t!]
    \centering
    \resizebox{\columnwidth}{!}{
    \begin{tabular}{cllllll}
    \toprule
         \multicolumn{1}{c|}{Method}  &\multicolumn{1}{c|}{Backbone}  &\multicolumn{1}{c}{Many} &\multicolumn{1}{c}{Medium} &\multicolumn{1}{c}{Few} &\multicolumn{1}{c}{All}\\ \midrule
         \multirow{1}*{OLTR~\cite{liu2019large}}  &\multicolumn{1}{|c|}{ResNet-32}  &\multicolumn{1}{c}{61.8} &\multicolumn{1}{c}{41.4} &\multicolumn{1}{c}{17.6} &\multicolumn{1}{c}{41.2} \\
         \multirow{1}*{LDAM+DRW~\cite{cao2019learning}}  &\multicolumn{1}{|c|}{ResNet-32}  &\multicolumn{1}{c}{61.5} &\multicolumn{1}{c}{41.7} &\multicolumn{1}{c}{20.2} &\multicolumn{1}{c}{42.0} \\
         \multirow{1}*{$\tau$-normalized~\cite{Kang2020Decoupling}}  &\multicolumn{1}{|c|}{ResNet-32}  &\multicolumn{1}{c}{65.7} &\multicolumn{1}{c}{43.6} &\multicolumn{1}{c}{17.3} &\multicolumn{1}{c}{43.2} \\
         \multirow{1}*{cRT~\cite{Kang2020Decoupling}}  &\multicolumn{1}{|c|}{ResNet-32}  &\multicolumn{1}{c}{64.0} &\multicolumn{1}{c}{44.8} &\multicolumn{1}{c}{18.1} &\multicolumn{1}{c}{43.3} \\
         \multirow{1}*{RIDE~\cite{wang2020long}}  &\multicolumn{1}{|c|}{ResNet-32}  &\multicolumn{1}{c}{69.3} &\multicolumn{1}{c}{49.3} &\multicolumn{1}{c}{26.0} &\multicolumn{1}{c}{49.1} \\
         \multirow{1}*{TADE~\cite{zhang2021test}}  &\multicolumn{1}{|c|}{ResNet-32}  &\multicolumn{1}{c}{65.4} &\multicolumn{1}{c}{49.3} &\multicolumn{1}{c}{29.3} &\multicolumn{1}{c}{49.8} \\  \midrule

         \multirow{4}*{\approach{}}  &\multicolumn{1}{|c|}{ResNet-50}  &\multicolumn{1}{c}{62.4} &\multicolumn{1}{c}{52.3} &\multicolumn{1}{c}{38.2} &\multicolumn{1}{c}{51.6 \textcolor{red}{(+1.8)}} \\
        ~&\multicolumn{1}{|c|}{ResNet-101}  &\multicolumn{1}{c}{69.5} &\multicolumn{1}{c}{59.3} &\multicolumn{1}{c}{47.1} &\multicolumn{1}{c}{59.2 \textcolor{red}{(+9.4)}} \\
        ~&\multicolumn{1}{|c|}{ViT-B/16}  &\multicolumn{1}{c}{\textbf{84.9}} &\multicolumn{1}{c}{\textbf{79.7}} &\multicolumn{1}{c}{\textbf{67.3}} &\multicolumn{1}{c}{\textbf{77.8} \textcolor{red}{\bf (+28.0)}} \\
        ~&\multicolumn{1}{|c|}{ResNet-50$\times$16}  &\multicolumn{1}{c}{74.6} &\multicolumn{1}{c}{62.8} &\multicolumn{1}{c}{52.0} &\multicolumn{1}{c}{63.7 \textcolor{red}{(+13.9)}} \\
    \bottomrule
    \end{tabular}
    }
    \caption{Long-tailed recognition performance comparison on CIFAR100-LT with an imbalance ratio of $100$. The red numbers represent the improvement of overall accuracy compared with the state-of-the-art performance (TADE with $49.8$\% overall accuracy).}
    \vspace{-15pt}
    \label{tab:cifar}
\end{table}

\noindent\textbf{CIFAR100-LT.~}
We also evaluate the models on CIFAR100-LT and show their performances in Table~\ref{tab:cifar}. As reported in the table, \approach{} outperforms the state-of-the-art expert-based ensemble methods RIDE~\cite{wang2021longtailed} and TADE~\cite{zhang2021test} by $+28.7\%$ and $+28.0\%$, respectively.

\subsection{Ablation Studies}
In this section, we conduct extensive ablation studies to validate the design choices of \approach{} from three aspects. We first explore how to best utilize vision-language backbone for finetuning. Moreover, we shows the effectiveness of linear adapter and how to make use of linear adapter for better performance. Finally, we demonstrate where and how to conduct data balancing.

\vspace{-6pt}
\subsubsection{Vision-Language Models}
We conduct ablations to demonstrate the effectiveness of vision-language backbones as introduced in Section~\ref{method}.

\noindent\textbf{The Effectiveness of Pretrained Weights.~}
% We utilize CLIP~\cite{radford2021learning} as the vision-language backbone for \approach{}. 
% CLIP is a vision-and-language model trained on 400M pairs of image and text descriptions collected from the internet. Due to the noisy data source, the impact of pretrained CLIP keeps unknown.
In Table~\ref{tab: pretrain}, we validate the effectiveness of pretrained CLIP weights compared with random initialized visual and language weights. All the four ablations are conducted on Phase A without data balancing for 50 epochs. The large gaps between random and pretrained CLIP initialization demonstrate the advantage of utilizing pretrained contrastive vision-language models.
% pretraining with contrastive learning shorten the training convergence and improve performance largely.
Besides, we find that visual encoder has much more influence than text encoder on the performance as random initialized vision encoder drops the accuracy close to zero. Note that poor performance of random initialization is primarily attributed to short training periods and pretrained vision-language weights fastening the convergence largely.

\begin{table}[t!]
    \centering
    \small
%    \resizebox{\textwidth}{!}{
    \begin{tabular}{lllllll}
    \toprule
         \multicolumn{1}{c|}{Vision}  &\multicolumn{1}{c|}{Language} &\multicolumn{1}{c}{Many} &\multicolumn{1}{c}{Medium} &\multicolumn{1}{c}{Few} &\multicolumn{1}{c}{All}\\ \midrule
         \multicolumn{1}{c|}{random}  &\multicolumn{1}{c|}{random} &\multicolumn{1}{c}{0.3} &\multicolumn{1}{c}{0.0} &\multicolumn{1}{c}{0.0} &\multicolumn{1}{c}{0.1}\\
         \multicolumn{1}{c|}{random}  &\multicolumn{1}{c|}{CLIP} &\multicolumn{1}{c}{0.3} &\multicolumn{1}{c}{0.0} &\multicolumn{1}{c}{0.0} &\multicolumn{1}{c}{0.1}\\
         \multicolumn{1}{c|}{CLIP}  &\multicolumn{1}{c|}{random} &\multicolumn{1}{c}{36.8} &\multicolumn{1}{c}{2.9} &\multicolumn{1}{c}{0.0} &\multicolumn{1}{c}{15.6}\\ 
         \multicolumn{1}{c|}{CLIP}  &\multicolumn{1}{c|}{CLIP} &\multicolumn{1}{c}{\bf 75.5} &\multicolumn{1}{c}{\bf 56.3} &\multicolumn{1}{c}{\bf 41.0} &\multicolumn{1}{c}{\textbf{61.6}}\\ 
    \bottomrule
    \end{tabular}
    \caption{Ablations of pretrained vision-language weights on ImageNet-LT dataset. \textit{CLIP} means using pre-trained weights as initialization and \textit{random} represents random initialization.}
    \label{tab: pretrain}
    \vspace{-6pt}
\end{table}

\begin{table}[t!]
    \centering
    \small
%    \resizebox{\textwidth}{!}{
    \begin{tabular}{lllllll}
    \toprule
         \multicolumn{1}{c|}{Vision}  &\multicolumn{1}{c|}{Language} &\multicolumn{1}{c}{Many} &\multicolumn{1}{c}{Medium} &\multicolumn{1}{c}{Few} &\multicolumn{1}{c}{All}\\ \midrule
         \multicolumn{1}{c|}{$\times$}  &\multicolumn{1}{c|}{$\times$} &\multicolumn{1}{c}{59.4} &\multicolumn{1}{c}{57.5} &\multicolumn{1}{c}{57.6} &\multicolumn{1}{c}{58.2}\\
         \multicolumn{1}{c|}{$\surd$}  &\multicolumn{1}{c|}{$\times$} &\multicolumn{1}{c}{70.4} &\multicolumn{1}{c}{\bf 65.4} &\multicolumn{1}{c}{\bf 58.0} &\multicolumn{1}{c}{\textbf{66.3}}\\
         \multicolumn{1}{c|}{$\times$}  &\multicolumn{1}{c|}{$\surd$} &\multicolumn{1}{c}{70.6} &\multicolumn{1}{c}{\bf 65.4} &\multicolumn{1}{c}{55.9} &\multicolumn{1}{c}{66.1}\\ 
         \multicolumn{1}{c|}{$\surd$}  &\multicolumn{1}{c|}{$\surd$} &\multicolumn{1}{c}{\bf 71.3} &\multicolumn{1}{c}{\bf 65.4} &\multicolumn{1}{c}{54.1} &\multicolumn{1}{c}{66.1}\\ 
    \bottomrule
    \end{tabular}
    \caption{Different methods of finetuning CLIP on ImageNet-LT. "$\surd$" means finetuning and "$\times$" means freezing the parameters of model. \textit{Vision} and \textit{Language} denotes the visual and text encoders of CLIP respectively. All models are finetuned for $50$ epochs.}
    \label{tab:finetune}
    \vspace{-13pt}
\end{table}

\noindent\textbf{Finetune the Vision-Language Model.~}
% As is introduced above, the vision-language models learn to align the vision and language representations with contrastive loss, attempting to utilize the rich textual information to facilitate the models' real comprehension of categories to recognize. 
% Different from the conventional pre-trained visual backbone like ResNet~\cite{he2016deep} or ViTs~\cite{dosovitskiy2021an} that are easily transferred to other tasks, how to adapt pretrained contrastive vision-language models for long-tailed recognition remains unexplored. We aim to empirically discover the optimal way to utilize contrastive vision-language models. 
To empirically discover how to utilize contrastive vision-language models, we probe the finetuning process by freezing the pre-trained image encoder and text encoder respectively. When both encoders are frozen, the model directly perform zero-shot predictions.
% outputs the classification results directly to do zero-shot prediction.
From Table~\ref{tab:finetune}, we can easily find the following pattern -- as more components are finetuned in CLIP, more accuracy improvement is obtained for \textit{many-shot} categories whereas more accuracy drop happens in \textit{few-shot} division. We hypothesize it is because the \textit{many-shot} classes dominate the visual feature space during finetuning. Therefore, for Phase A, it is necessary to adapt CLIP on specific long-tailed dataset as much as possible, and we choose to finetune both the vision and language branches of CLIP.
%Therefore, we can conclude that finetuning on visual branch alone helps pretrained vision-language model refine its representations for all three subsets in specific long-tailed dataset.
% modify the weight space to fit the decision boundary. 
% We hypothesize that the finetuning process prompts the pre-trained CLIP model to adapt the unbiased decision boundary into the targeted distribution of long-tailed feature embeddings \shijie{don't understand this sentence very well}, as \textit{many-shot} classes occupy the dominant feature embedding space of the long-tailed dataset distribution. 

\begin{figure}[t!]
   \centering
   \vspace{-10pt}
   \hspace{-2mm}
    \subcaptionbox{ImageNet-LT}{\includegraphics[width = 0.5\linewidth]{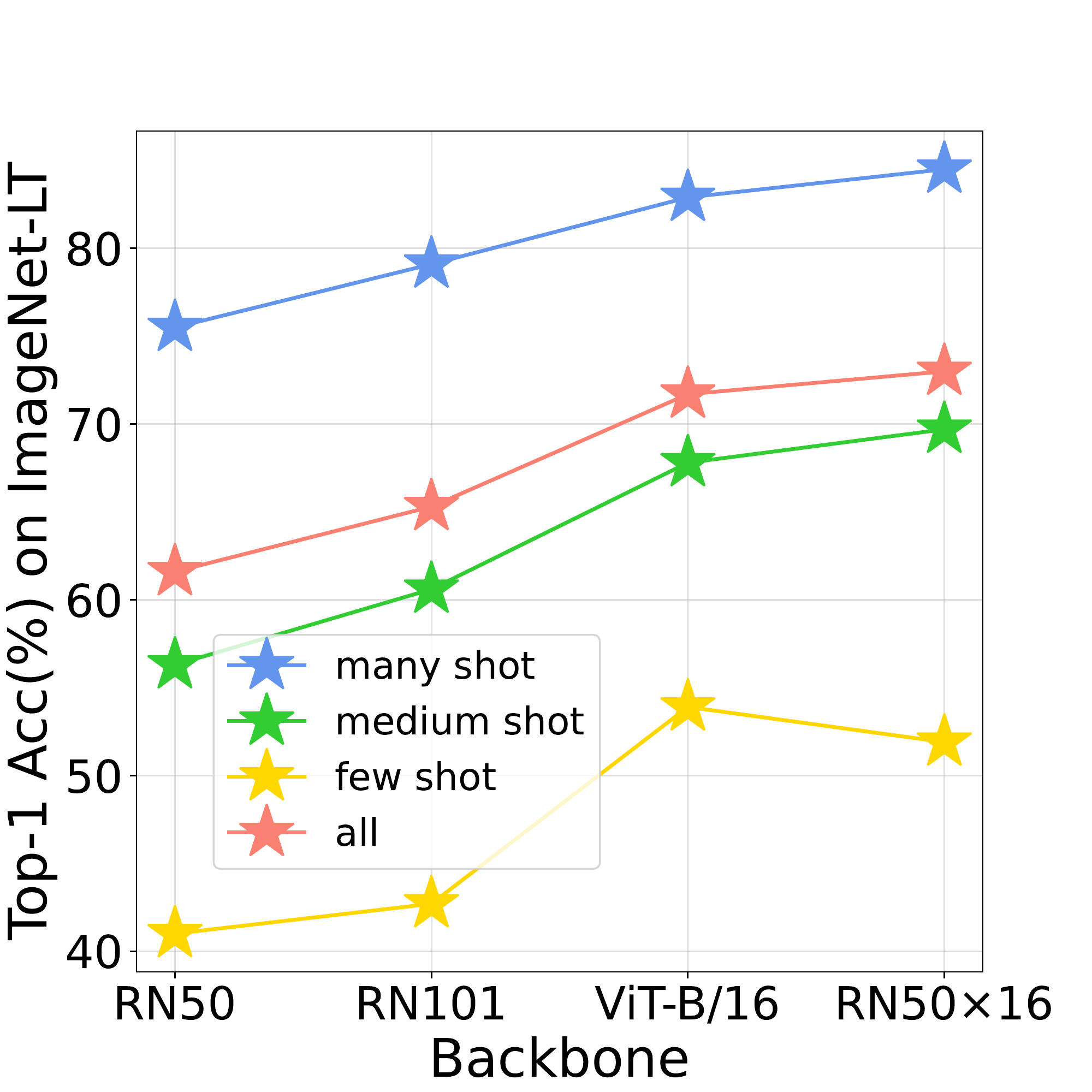}}\hfill
    %\subcaption{Weight distribution}
    \hspace{-8mm}
	\subcaptionbox{Places-LT.}{\includegraphics[width = 0.5\linewidth]{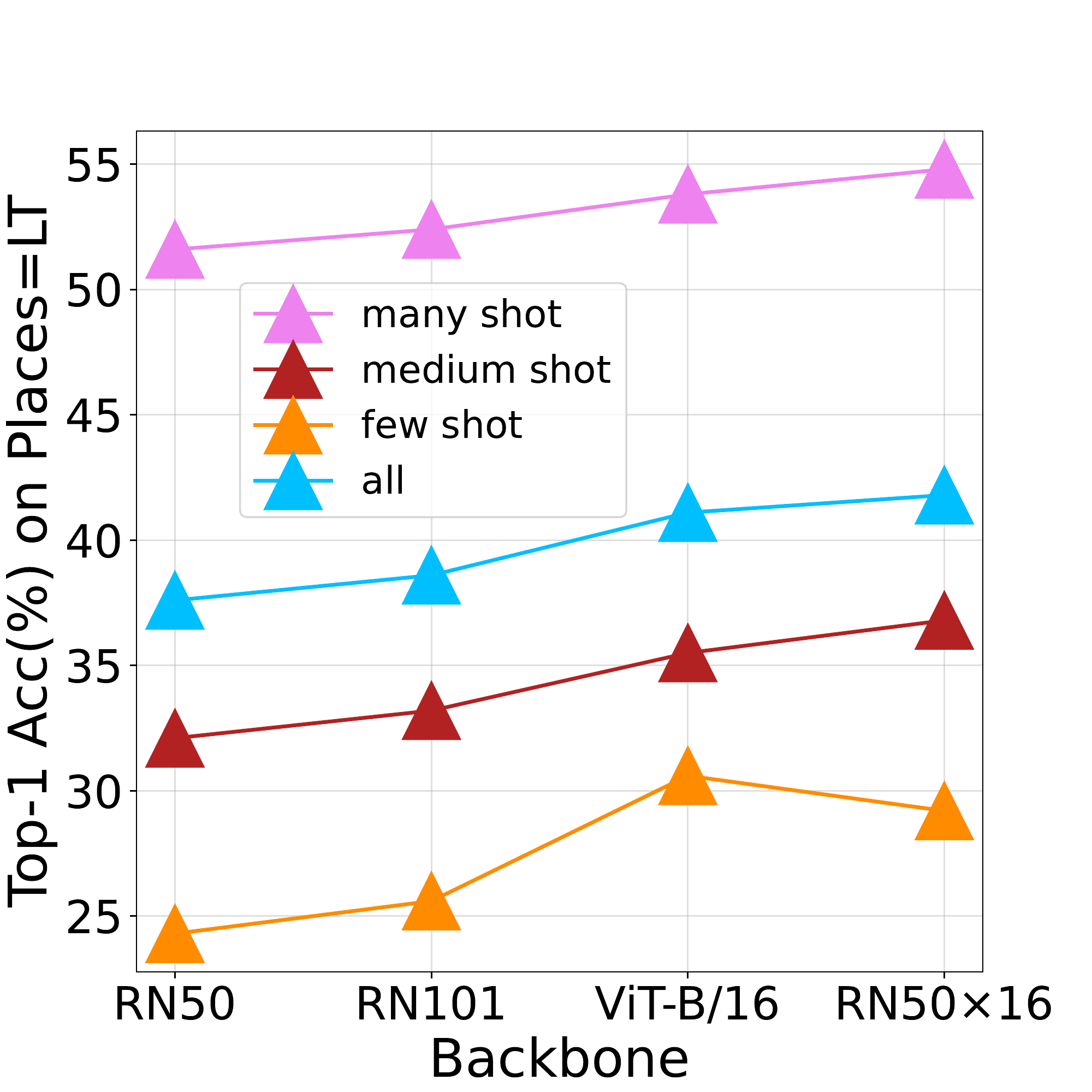}}
	%\subcaption{Weight distribution}
	\caption{Comparisons between several visual backbones for ImageNet-LT (left) and Places-LT (right).}
	\vspace{-15pt}
	\label{fig:backbone}
\end{figure}

\noindent\textbf{Visual Backbones.}
% Different visual backbones of CLIP model exert different influence on final performance of \approach{} considered the same text encoder. 
We try CLIP with different visual backbones to explore its influence on final performance of \approach{}. We report the Phase A results of different backbones in Figure~\ref{fig:backbone} on both ImageNet-LT and Places-LT benchmarks. When the visual backbone becomes deeper and larger, the finetuned performance is also gradually improved for \textit{all}, \textit{many-shot}, and \textit{medium-shot} categories. Surprisingly, the Vision Transformer structure~\cite{dosovitskiy2021an} achieves the best accuracy in \textit{few-shot} subset, probably owing to multi-head self-attention mechanism's ability in capturing minor features.
\vspace{-10pt}
% We speculate the multi-head self-attention mechanism of ViTs dynamically update attention map based on global visual features, and combine the embeddings from different heads space, thus capturing abundant minor features, which is fundamental to \textit{few-shot} categories considering insufficient instances.

% \noindent\textbf{Text prompt.}
% We reveal the impact of text prompt choices and prompt ensemble when finetuing the Vision and Language model specifically in Appendix.

% \begin{figure}[ht!]
% \centering
% \includegraphics[width=1.05\linewidth,trim={0cm 0cm 0cm 0cm}]{latex/images/backbone.pdf}
% \vspace*{-13pt}
% \caption{Comparisons between several vision backbones for ImageNet-LT (left) and Places-LT (right).}
% \vspace*{-13pt}
% \label{fig:backbone}
% \end{figure}

\subsubsection{Linear Adapter}
\label{subsec:adapter}
% In Phase B of \approach{}, an auxiliary linear adapter is adopted to further finetune the visual backbone for better performance of \textit{few-shot} classes with negligible additional cost. 
We validate the effectiveness of adopting linear adapter in Phase B and explore the key factors that determines the performance of linear adapter.

\begin{figure}[t!]
\centering
\includegraphics[width=0.8\linewidth,trim={0cm 0cm 0cm 0cm}]{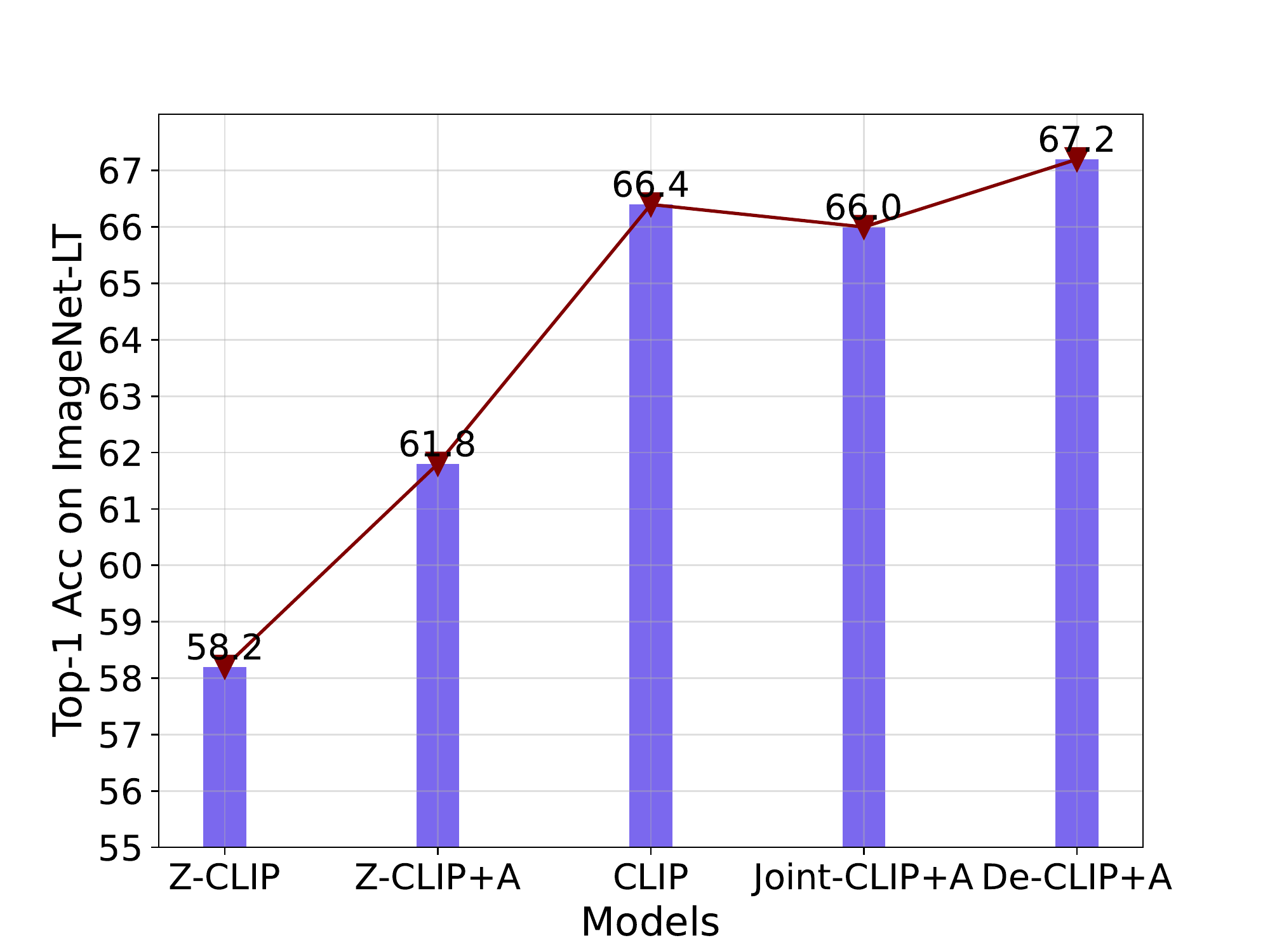}
\caption{Ablations of effectiveness of Linear Adapter and decouple finetuning. \textit{Z-CLIP}: zero-shot CLIP model; \textit{Z-CLIP+A}: finetune adapter based on zero-shot CLIP; \textit{CLIP}: directly finetune CLIP; \textit{Joint-CLIP+A}: jointly finetune CLIP and adapter; \textit{De-CLIP+A}: the \approach{} style which decouples the CLIP and adapter finetuning into two phases.}
\vspace*{-8pt}
\label{fig:decouple}
\end{figure}

\noindent\textbf{The Effectiveness of Linear Adapter.~}
We design ablations to demonstrate the influence of linear adapter. First, we freeze the parameters of CLIP and finetune the linear adapter for $10$ epochs to mix the original zero-shot visual embedding with the corresponding finetuned visual features via residual connection. As illustrated in Figure~\ref{fig:decouple}, the simple $10$-epoch training improves the performance from $58.2\%$ to $61.8\%$ even without data balancing. 
Moreover, we finetune both the visual and language encoders of CLIP for $50$ epochs and then finetune the linear adapter for another $10$ epochs with CLIP parameters fixed. Compared with an equal $60$-epoch training scheme of finetuning the visual and language encoder of the CLIP, an extra $10$-epoch finetuning of linear adapter based on $50$-epoch finetuning of CLIP backbone can further boost the top-1 accuracy from $66.4\%$ to $67.2\%$.
%We visualize the effectiveness of linear adapter to prompt for the classification of the model.

\begin{table}[t!]
    \centering
    \small
    \resizebox{\columnwidth}{!}{
    \begin{tabular}{lllllll}
    \toprule
         \multicolumn{1}{c|}{V-Adapter}  &\multicolumn{1}{c|}{L-Adapter} &\multicolumn{1}{c}{Many} &\multicolumn{1}{c}{Medium} &\multicolumn{1}{c}{Few} &\multicolumn{1}{c}{All}\\ \midrule
         \multicolumn{1}{c|}{$\surd$}  &\multicolumn{1}{c|}{$\times$} &\multicolumn{1}{c}{\bf 71.0} &\multicolumn{1}{c}{\bf 66.3} &\multicolumn{1}{c}{\bf59.5} &\multicolumn{1}{c}{\textbf{67.2}}\\
         \multicolumn{1}{c|}{$\times$}  &\multicolumn{1}{c|}{$\surd$} &\multicolumn{1}{c}{\bf 71.0} &\multicolumn{1}{c}{66.2} &\multicolumn{1}{c}{59.0} &\multicolumn{1}{c}{67.0}\\ 
         \multicolumn{1}{c|}{$\surd$}  &\multicolumn{1}{c|}{$\surd$} &\multicolumn{1}{c}{70.6} &\multicolumn{1}{c}{66.2} &\multicolumn{1}{c}{58.4} &\multicolumn{1}{c}{66.8}\\ 
    \bottomrule
    \end{tabular}}
    \caption{Variants of linear adapter. \textit{V-Adapter} and \textit{L-Adapter} represents using linear adapter layer to adapt visual and language encoders respectively. All results are trained on ImageNet-LT for $10$ epochs.}
    \vspace*{-15pt}
    \label{tab:where adapt}
\end{table}

\noindent\textbf{Should the Finetuning of CLIP and Linear Adapter be Decoupled?~}
As mentioned in Section~\ref{method} and illustrated in Figure~\ref{fig:1}, we decouple the training process into two phases -- in Phase A, we train both the vision and language encoder of CLIP based on pre-trained weights; in Phase B, we freeze the parameters of visual and language encoders while only finetuning the linear adapter. An alternative scheme is to jointly train the CLIP and linear adapter rather than decoupling the training processes. According to Figure~\ref{fig:decouple}, joint training of CLIP and linear adapter (\emph{Joint-CLIP+A}) leads to a $0.4\%$ accuracy drop compared with directly finetuning CLIP without adapter (\emph{CLIP}). In contrast, the decoupled training of CLIP and linear adapter (\emph{De-CLIP+A}) can largely boost the accuracy from $66.0\%$ to $67.2\%$ and the ascent mainly comes from tail classes, which is up to $6.0\%$.
We visualize the joint and decoupling training schemes using t-SNE~\cite{van2008visualizing} and present the results in the supplementary. Compared with joint training, decoupled training better separates the tail-class feature embeddings from head-classes. This demonstrates that the proposed decoupled training of vision-language model and adapter is effective to handle long-tailed distribution.

\noindent\textbf{Variants of Linear Adapter.~}
Since CLIP has dual encoders, the auxiliary linear adapter could be added to either or both of the two branches. As reported in Table~\ref{tab:where adapt}, we try linear adapter for adapting visual and language encoders respectively.
% We set the residual factor $\lambda$ to $0.2$, which represents the coefficients of residual connection.
% We perform experiments based on finetuned CLIP from Phase A and continuously train the linear adapter for $10$ epochs in Phase B. 
From the table, we can find that applying the linear adapter to the visual branch of CLIP achieves the best overall performance and is the optimal choice.
% of adapted visual/language features and original visual/language features, respectively. 

\begin{figure}[t]
\centering
\includegraphics[width=0.8\linewidth,trim={0cm 0cm 0cm 0cm}]{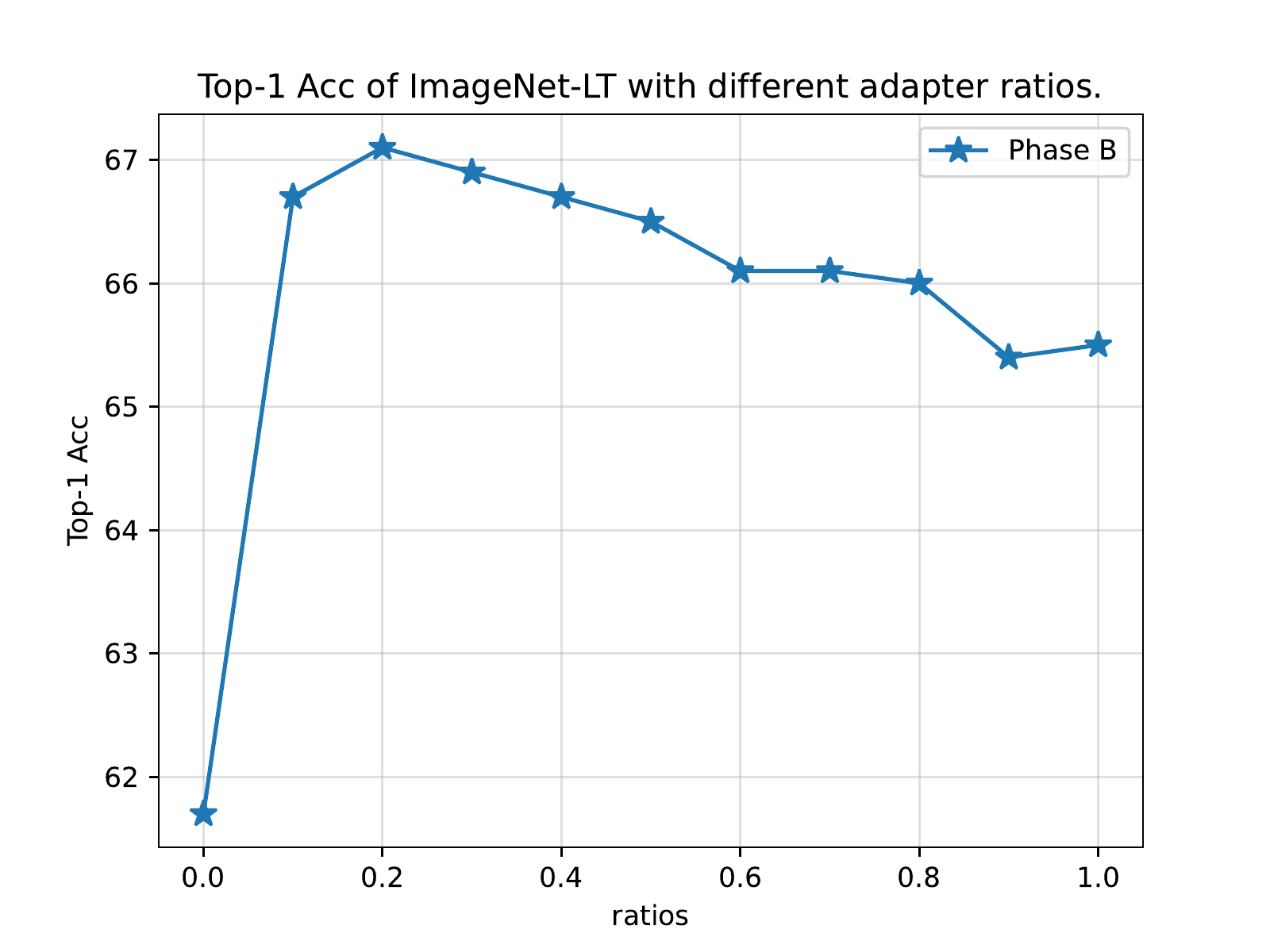}
\vspace*{-13pt}
\caption{The top-1 accuracy on ImageNet-LT with different values of residual factor $\lambda$ in Phase B of \approach{}.}
\vspace*{-13pt}
\label{fig:ratio}
\end{figure}

\noindent\textbf{Hyperparameters of the Linear Adapter.~}
Moreover, we explore the influence of linear adapter's residual factor $\lambda$. $\lambda$ determines the importance of new knowledge obtained from finetuning the linear adapter.
Note that when $\lambda$ equals to $1.0$, the classification is fully determined by the adapted image features. We explore different values of $\lambda$ from $0.0$ to $1.0$ and conduct the ablations of Phase B finetuning on ImageNet-LT.
% The ablations of different ratios are conducted in Phase B with finetuned CLIP model fixed on the dataset of ImageNet-LT.
As shown in Figure~\ref{fig:ratio}, the best performance of linear adapter can be obtained when $\lambda$ is around $0.2$, with a top-1 accuracy of $67.1\%$ on ImageNet-LT. The empirical results reveal the knowledge of finetuned CLIP is already good enough to handle most cases, a slight and balanced adaptation would further improve the performance.

\vspace*{-13pt}

\subsubsection{Balancing Methods}
%Balancing methods are important and effective in tackling the long-tailed distribution tasks. 
Balancing methods can alleviate the severe performance degradation due to class imbalance. 
% As introduced in Section~\ref{sec:related}, many balancing methods have been explored to solve the problem. 
In this section, we explore different balancing methods for \approach{} to reveal two significant problems: 1) where to utilize balancing methods, and 2) which balancing methods to apply. 
%Note in this paper we pay more attention to balance sampling strategies to unify the structure of \approach{}, some other effective strategies like loss re-weighting, $\tau$-normalization would be omitted as  

\noindent\textbf{Where to balance.~}
% As discussed above, \approach{} first balances the long-tailed data in Phase B and then trains a single linear adapter for finetuning.
% We compare this training scheme with other alternatives such as balancing the data distribution in Phase A and in both Phase A and Phase B.
Here, we compare balancing the long-tailed data distribution on either or both of two phases.
The experiments are performed on ImageNet-LT and Places-LT datasets with ResNet-50-backboned CLIP.
%本身就有很均衡的知识，但是finetune会破坏本身的知识均衡，但学习到了特征，利用linear adapter重新均衡新学习到的知识
%The experimental results in Table~\ref{tab:balance} suggest that in Phase A, a balancing process is unnecessary as sacrificing many-shot categories would hinder the vision-language representation learning. Trained with image-text pairs rather than fixed class labels, CLIP is capable of dealing with the imbalanced data and shows great potential of predicting the few-shot categories. Therefore, the finetuning process of CLIP should not be disrupted by all kinds of balancing strategies, as the main goal of Phase A is to shift the internal decision domain of CLIP from web-sourced noisy data to specific long-tailed task datasets. 
% As is known, \textit{many-shot} categories make up dominant features due to \shijie{what is head distribution?} head distribution.
As mentioned earlier, \textit{many-shot} categories dominate the feature space of long-tailed distribution. The performance drops of \textit{many-shot} categories on both datasets, as reported in Table~\ref{tab:balance}, suggest that balancing during Phase A tends to sacrifice \textit{many-shot} representations. Since Phase A is mainly designed for updating representations on a new domain, we thereby abandon Phase-A data balancing.
% Performance drops reported in Table~\ref{tab:balance} of \textit{many-shot} categories on both datasets when balancing during Phase A suggest balancing tends to sacrifice \textit{many-shot} representations. 
% As Phase A is designed for updating representations on a new domain, we abandon Phase A balancing thereby.
% A richer feature could help filter the noisy and confusing embeddings.
% by extracting as rich features as possible to filter the noisy and confusing embeddings. 
% In this process, the balance between different-shots categories should be avoided to some extent, however, the specific features of target domain have been successfully obtained as the \textit{many-shot} accuracy raises a lot. 
When applying balancing strategies to Phase B alone, \approach{} can achieve a more balanced performance for different shots and improve the overall top-1 accuracy thanks to the rich features learned from Phase A.
%由于不是在固定类别上而是在image-text pair上训练的，CLIP本身就有很强的应对长尾分布的能力，因此，在对clip finetune时也不需要考虑均衡问题，只需要最大程度上地适应长尾分布的特征，随后在第二阶段通过一个简单的linear adapter配合balance方法 就可以使得在各种shot上的表现重新趋于均衡，并且整体表现也会增加，ImageNet-LT上增加0.3个点， Places-LT上增加了1个点。

\begin{table}[t!]
    \centering
    \resizebox{\columnwidth}{!}{
    \begin{tabular}{cllllll}
    \toprule
         \multirow{2}*{Dataset}  &\multicolumn{2}{|c|}{Balance}  &\multirow{2}*{Many} &\multirow{2}*{Medium} &\multirow{2}*{Few} &\multirow{2}*{All}\\ 
         ~&\multicolumn{1}{|c}{Phase A} &\multicolumn{1}{c|}{Phase B} & & & & \\
         \midrule
         \multirow{4}*{ImageNet-LT}  &\multicolumn{1}{|c}{$\times$} &\multicolumn{1}{c|}{$\times$} &\multicolumn{1}{c}{\bf 77.3} &\multicolumn{1}{c}{57.4} &\multicolumn{1}{c}{39.0} &\multicolumn{1}{c}{62.6} \\
         ~  &\multicolumn{1}{|c}{$\surd$} &\multicolumn{1}{c|}{$\times$} &\multicolumn{1}{c}{76.6} &\multicolumn{1}{c}{58.4} &\multicolumn{1}{c}{42.7} &\multicolumn{1}{c}{63.3} \\
         ~  &\multicolumn{1}{|c}{$\surd$} &\multicolumn{1}{c|}{$\surd$} &\multicolumn{1}{c}{70.7} &\multicolumn{1}{c}{66.2} &\multicolumn{1}{c}{58.5} &\multicolumn{1}{c}{66.9} \\
         ~  &\multicolumn{1}{|c}{$\times$} &\multicolumn{1}{c|}{$\surd$} &\multicolumn{1}{c}{71.0} &\multicolumn{1}{c}{\bf 66.3} &\multicolumn{1}{c}{\bf 59.5} &\multicolumn{1}{c}{\textbf{67.2}} \\  \midrule

         \multirow{4}*{Places-LT} &\multicolumn{1}{|c}{$\times$} &\multicolumn{1}{c|}{$\times$} &\multicolumn{1}{c}{\bf 52.7} &\multicolumn{1}{c}{32.9} &\multicolumn{1}{c}{23.4} &\multicolumn{1}{c}{38.2} \\
         ~  &\multicolumn{1}{|c}{$\surd$} &\multicolumn{1}{c|}{$\times$} &\multicolumn{1}{c}{51.3} &\multicolumn{1}{c}{33.2} &\multicolumn{1}{c}{25.5} &\multicolumn{1}{c}{38.2} \\
         ~  &\multicolumn{1}{|c}{$\surd$} &\multicolumn{1}{c|}{$\surd$} &\multicolumn{1}{c}{44.6} &\multicolumn{1}{c}{46.7} &\multicolumn{1}{c}{44.1} &\multicolumn{1}{c}{45.5} \\
         ~  &\multicolumn{1}{|c}{$\times$} &\multicolumn{1}{c|}{$\surd$} &\multicolumn{1}{c}{46.7} &\multicolumn{1}{c}{\bf 48.0} &\multicolumn{1}{c}{\bf 42.7} &\multicolumn{1}{c}{\textbf{46.5}} \\
    \bottomrule
    \end{tabular}
    }
    \caption{Where to employ balance strategies ablations. On both ImageNet-LT and Places-LT, balance only in Phase B make \approach{} perform the best.}
    \vspace{-3pt}
    \label{tab:balance}
\end{table}

\begin{table}[t!]
    \centering
    \footnotesize
    \resizebox{0.92\columnwidth}{!}{
    \begin{tabular}{llllll}
    \toprule
         \multicolumn{1}{c|}{Balance Methods}  &\multicolumn{1}{c}{Many} &\multicolumn{1}{c}{Medium} &\multicolumn{1}{c}{Few} &\multicolumn{1}{c}{All}\\ \midrule
         \multicolumn{1}{c|}{Class-balanced}   &\multicolumn{1}{c}{71.0} &\multicolumn{1}{c}{\textbf{66.3}} &\multicolumn{1}{c}{\textbf{59.5}} &\multicolumn{1}{c}{\textbf{67.2}}\\
         \multicolumn{1}{c|}{Square-root}   &\multicolumn{1}{c}{\textbf{75.2}} &\multicolumn{1}{c}{62.8} &\multicolumn{1}{c}{50.9} &\multicolumn{1}{c}{66.0}\\ 
         \multicolumn{1}{c|}{Mix-balanced}   &\multicolumn{1}{c}{72.6} &\multicolumn{1}{c}{64.9} &\multicolumn{1}{c}{59.1} &\multicolumn{1}{c}{67.1}\\ 
    \bottomrule
    \end{tabular}}
    \caption{Comparison of different balanced sampling strategies on ImageNet-LT.}
    \vspace*{-13pt}
    \label{tab:sampling}
\end{table}

\noindent\textbf{How to balance.~}
%sampling
Furthermore, we explore different sampling strategies including class-balanced sampling, square-root sampling and mix-balanced sampling for Phase B.
% We compare the performance achieved by class-balanced sampling, square-root sampling and Mix-balanced sampling in Phase B. 
Class-balanced sampling samples the categories from original dataset in equal probability rather than the natural instance-balanced sampling which selects instances regardless of classes. The process can be decoupled into two steps -- first selecting classes equally from the list of categories and then randomly sampling a data point from the selected class. Square-root sampling~\cite{mahajan2018exploring} first computes the square-root of the number of head classes, then re-normalize and conduct sampling according to the resulting distribution. Mix-balanced sampling combines the instance-balanced sampling and class-balanced sampling, thus takes advantage of both strategies to avoid overfitting at early epochs and underfitting at late epochs. Motivated by the~\cite{Kang2020Decoupling}, we adopt a soft version of mix-balanced sampling to dynamically interpolates between instance-balanced sampling and class-balanced sampling as learning progresses. As shown in Table~\ref{tab:sampling}, class-balanced sampling can best benefit \emph{medium-shot} and \emph{few-shot} categories. Thus we adopt class-balanced sampling as the balancing method of \approach{}.

\section{Conclusion}
In this paper, we proposed \approach{} which tackles long-tailed recognition by leveraging contrastive vision-language models. 
We decouple \approach{} into two phases for training with long-tailed and balanced samples respectively. We first continue pretraining with contrastive loss to fully utilize abundant data to update visual-language representation on specific domains. After that, we employ an auxiliary linear adapter to refine the visual representation of tail classes. 
%As a result, \approach{} beats competitive baselines by a decent margin and achieves the new state-of-the-art performances on three benchmarks. 
We hope our simple \approach{} baseline could stimulate more future researches on exploring vision-language models for long-tailed recognition.

% \noindent\textbf{Limitations and Potential Negative Societal Impact.~}
% One limitation of this research is that finetuning language models is computationally expensive when the number of categories are huge, as enlarging the dimensions of Transformers~\cite{vaswani2017attention} will raise the computation cost quadratically. Thereby, it brings about potential societal impacts of carbon emission. 
%Also, we should be aware of the power of neural networks and prevent them from being used for harmful applications.
% The work is a development of your celebrated 1968 paper entitled ``Zero-g frobnication: How being the only people in the world with access to the Apollo lander source code makes us a wow at parties'', by Zeus \etal. \eg

% \newpage
%%%%%%%%% REFERENCES
{\small
\bibliographystyle{ieee_fullname}
\bibliography{long-tail}
}

%\newpage

\begin{center}
    \Large
    \textbf{Appendices}
    \\[10pt]
\end{center}

\appendix
This supplementary file includes more details and results of \approach{} that were not contained in the main manuscript due to the limited paper size. We start with elaborating the two-phase training of \approach{} in Appendix~\ref{algorithm}. Then we provide the zero-shot long-tailed recognition performance of the vision-language model CLIP in three benchmark datasets (Appendix~\ref{zeroshot-per}). Moreover, we perform ablations of different text prompts in Appendix~\ref{text}. Appendix~\ref{visualization} shows visualization results of the decoupled training of CLIP and linear adapter as discussed in Sec.~\ref{subsec:adapter} in main manuscript.

\section{Algorithm}
\label{algorithm}
We provide the pseudo code of training \approach{} as shown in Algorithm~\ref{pseudo}. In phase A, we keep training the vision-language model, updating parameters of both visual and language encoders. Afterward, in phase B, we train a single linear adapter with vision-language model frozen to adapt visual features in a balanced way. 
\begin{algorithm}
\caption{Two-phases training of \approach{}}
\label{pseudo}
\begin{algorithmic}
\Require Training samples $\{(\bm{I},y)\}$, %text sequence $T$, 
visual and language encoder $\mathcal{V}_{\mathrm{enc}}$, $\mathcal{L}_{\mathrm{enc}}$, linear adapter $\mathcal{LA}$
%phase A, B training period $N_A$, $N_B$
\State \textbf{Initialize} $\mathcal{V}_{\mathrm{enc}}$, $\mathcal{L}_{\mathrm{enc}}$ with web-data pretrained parameters $\Theta_{v}$  and $\Theta_{l}$
\For{epoch $ = 1,..., N_A$}   \Comment{Phase A}
\For{minibatch $B \in \{(\bm{I},y)\}$} 
\State $\bm{f}_{v} \gets \mathcal{V}_{\mathrm{enc}}(\bm{I}) \in \mathbb{R}^{d_v}$
\State $\bm{T} \gets tokenize(y)$
\State $\bm{f}_{l} \gets \mathcal{L}_{\mathrm{enc}}(\bm{T}) \in \mathbb{R}^{d_l}$
\State Project into embedding space $\mathbf{u}, \mathbf{v}$ as Eq.(\ref{eq:joint})
% \State $\mathbf{v} = \frac{\mathbf{W}_{v}^{\top}\bm{f}_v}{\|\mathbf{W}_{v}^{\top}\bm{f}_v\|}, \ \ \  \mathbf{u} = \frac{\mathbf{W}_{l}^{\top}\bm{f}_l}{\|\mathbf{W}_{l}^{\top}\bm{f}_l\|}$
\State $p_{i} \gets \frac{\exp \left(\mathbf{v}^{\top} \mathbf{u}_i\right) / \tau}{\sum_{j=1}^{K} \exp \left(\mathbf{v}^{\top} \mathbf{u}_j\right) / \tau}$
\State Update $\Theta_{v}$  and $\Theta_{l}$
% \State Compute the class probability as Eq.\textcolor{red}{4}
\EndFor
\EndFor

\vspace{4pt}
\State \textbf{Initialize} $\Theta_{LA}$ randomly for $\mathcal{LA}$  \Comment{Phase B}
\State \textbf{Freeze} $\Theta_{v}$  and $\Theta_{l}$  
\For{epoch $ = 1,..., N_B$}  
\For{minibatch $B \in \{Balanced(\bm{I},y)\}$}
\State $\bm{f}_{v} \gets \lambda \mathcal{LA}(\mathcal{V}_{\mathrm{enc}}(\bm{I})) + (1-\lambda)\mathcal{V}_{\mathrm{enc}}(\bm{I}) \in \mathbb{R}^{d_v}$
\State $\bm{T} \gets tokenize(y)$
\State $\bm{f}_{l} \gets \mathcal{L}_{\mathrm{enc}}(\bm{T}) \in \mathbb{R}^{d_l}$
\State Project into embedding space $\mathbf{u}, \mathbf{v}$ as Eq.(\ref{eq:joint})
% \State $\mathbf{v} = \frac{\mathbf{W}_{v}^{\top}\bm{f}_v}{\|\mathbf{W}_{v}^{\top}\bm{f}_v\|}, \ \ \  \mathbf{u} = \frac{\mathbf{W}_{l}^{\top}\bm{f}_l}{\|\mathbf{W}_{l}^{\top}\bm{f}_l\|}$
\State $p_{i} \gets \frac{\exp \left(\mathbf{v}^{\top} \mathbf{u}_i\right) / \tau}{\sum_{j=1}^{K} \exp \left(\mathbf{v}^{\top} \mathbf{u}_j\right) / \tau}$
\State Update $\Theta_{LA}$
% \State Compute the class probability as Eq.\textcolor{red}{4}
\EndFor
\EndFor
\end{algorithmic}
\end{algorithm}

\section{Zero-shot Performance}
\label{zeroshot-per}
The zero-shot long-tailed recognition performances on three benchmark datasets are presented in Table~\ref{zeroshot}. The red numbers show how much \approach{} has improved compared with zero-shot CLIP results. It illustrates the training scheme of \approach{} is effective as raising the initial vision-language model by a large margin ($+12.2\%$ maximally).

\begin{table*}[t!]
	\centering
% 	\tiny
    % \resizebox{\textwidth}{!}{
    	\begin{tabular}{c|cc|cc|cc}
    		\hline
    		\multirow{2}*{Visual Backbone} & \multicolumn{2}{c|}{ImageNet-LT} & \multicolumn{2}{c|}{Places-LT} & \multicolumn{2}{c}{CIFAR100-LT}  \\
    		~ & zero-shot & \approach{} &zero-shot & \approach{} & zero-shot  & \approach{} \\
    		\hline
            ResNet-50 & 58.2  & 67.2 (\textcolor{red}{+9}) &  35.3 & 46.5 (\textcolor{red}{+11.2}) & 40.2 & 51.6 (\textcolor{red}{+11.4}) \\
            
            ResNet-101 & 61.2 & 70.5 (\textcolor{red}{+9.3}) &  36.2 & 47.9 (\textcolor{red}{+11.7}) & 47.8 & 59.2 (\textcolor{red}{+11.4}) \\

            ViT-B/16 & 66.7 & 75.7 (\textcolor{red}{+9}) &  37.8 & 49.5 (\textcolor{red}{+11.7}) & 66.4 & 77.8 (\textcolor{red}{+11.4}) \\
    	
            ResNet-50$\times 16$ & 69.0 & 76.5 (\textcolor{red}{+7.5}) &  37.1 & 49.3 (\textcolor{red}{+12.2}) & 52.9 & 63.7 (\textcolor{red}{+10.8}) \\
    		\hline
    	\end{tabular}
    % 	}
    \caption{Top-1 accuracy of zero-shot CLIP and \approach{}-training.}
    \label{zeroshot}
\vspace{-0.3cm}
\end{table*}

\section{Text Prompting}
\label{text}

\begin{table}[t!]
    \centering
    % \footnotesize
    \resizebox{0.92\columnwidth}{!}{
    \begin{tabular}{llllll}
    \toprule
         \multicolumn{1}{c|}{Text Prompts}  &\multicolumn{1}{c}{Many} &\multicolumn{1}{c}{Medium} &\multicolumn{1}{c}{Few} &\multicolumn{1}{c}{All}\\ \midrule
         \multicolumn{1}{c|}{Single prompt}   &\multicolumn{1}{c}{\textbf{71.0}} &\multicolumn{1}{c}{\textbf{66.3}} &\multicolumn{1}{c}{59.5} &\multicolumn{1}{c}{\textbf{67.2}}\\
         \multicolumn{1}{c|}{Random single prompt}   &\multicolumn{1}{c}{70.5} &\multicolumn{1}{c}{66.1} &\multicolumn{1}{c}{\textbf{59.8}} &\multicolumn{1}{c}{66.9}\\ 
         \multicolumn{1}{c|}{Ensemble prompts}   &\multicolumn{1}{c}{70.5} &\multicolumn{1}{c}{66.0} &\multicolumn{1}{c}{59.5} &\multicolumn{1}{c}{66.8}\\ 
    \bottomrule
    \end{tabular}}
    \caption{Comparison of different balanced sampling strategies on ImageNet-LT.}
    \vspace*{-6pt}
    \label{prompt}
\end{table}

Prompt engineering is initially proposed for knowledge probing in large pretrained language models~\cite{petroni2019language, shin2020eliciting, li2021prefix, jiang2020can}. Prompting is adding extra instructions to task inputs to generate specific outputs from pretrained language model. In this paper, we utilize manually designed prompts following CLIP~\cite{radford2021learning}. Specifically, a prompt template like \textit{a photo of a} \{\texttt{CLASS}\} is adopted to all experiments reported in main manuscript. However, CLIP~\cite{radford2021learning} claims that ensembling several classifiers using different hand-crafted prompts as follows can improve the performance of zero-shot tasks.

\begin{itemize}
    \item itap of a \{\texttt{CLASS}\}.
    \item a bad photo of the \{\texttt{CLASS}\}.
    \item a origami \{\texttt{CLASS}\}.
    \item a photo of the large \{\texttt{CLASS}\}.
    \item a \{\texttt{CLASS}\} in the video game.
    \item art of the \{\texttt{CLASS}\}.
    \item a photo of the small \{\texttt{CLASS}\}.
\end{itemize}

Therefore, we perform ablations on ensembling the multiple prompts and randomly choosing one of them for language model training. Ablations are conducted on ImageNet-LT benchmark~\cite{liu2019large} with ResNet-50 visual backbone. Results are reported in Table~\ref{prompt}. Surprisingly, prompts ensembling decays performance from $67.2\%$ to $66.8\%$ rather than raising accuracy. Randomly choosing a template from above seven prompts also results in performance drop, by $0.3\%$ overall accuracy. We hypothesize different prompting templates from multiple views may confuse the pretrained language model finetuning, which is different from zero-shot task. Thus, we choose a single template \textit{a photo of a} \{\texttt{CLASS}\} in all our ablations.

\section{Visualization}
\label{visualization}
\begin{figure}[t!]
   \centering
   \vspace{-10pt}
   \hspace{-2mm}
    \subcaptionbox{Decoupled training.}{\includegraphics[width = 1.0\linewidth]{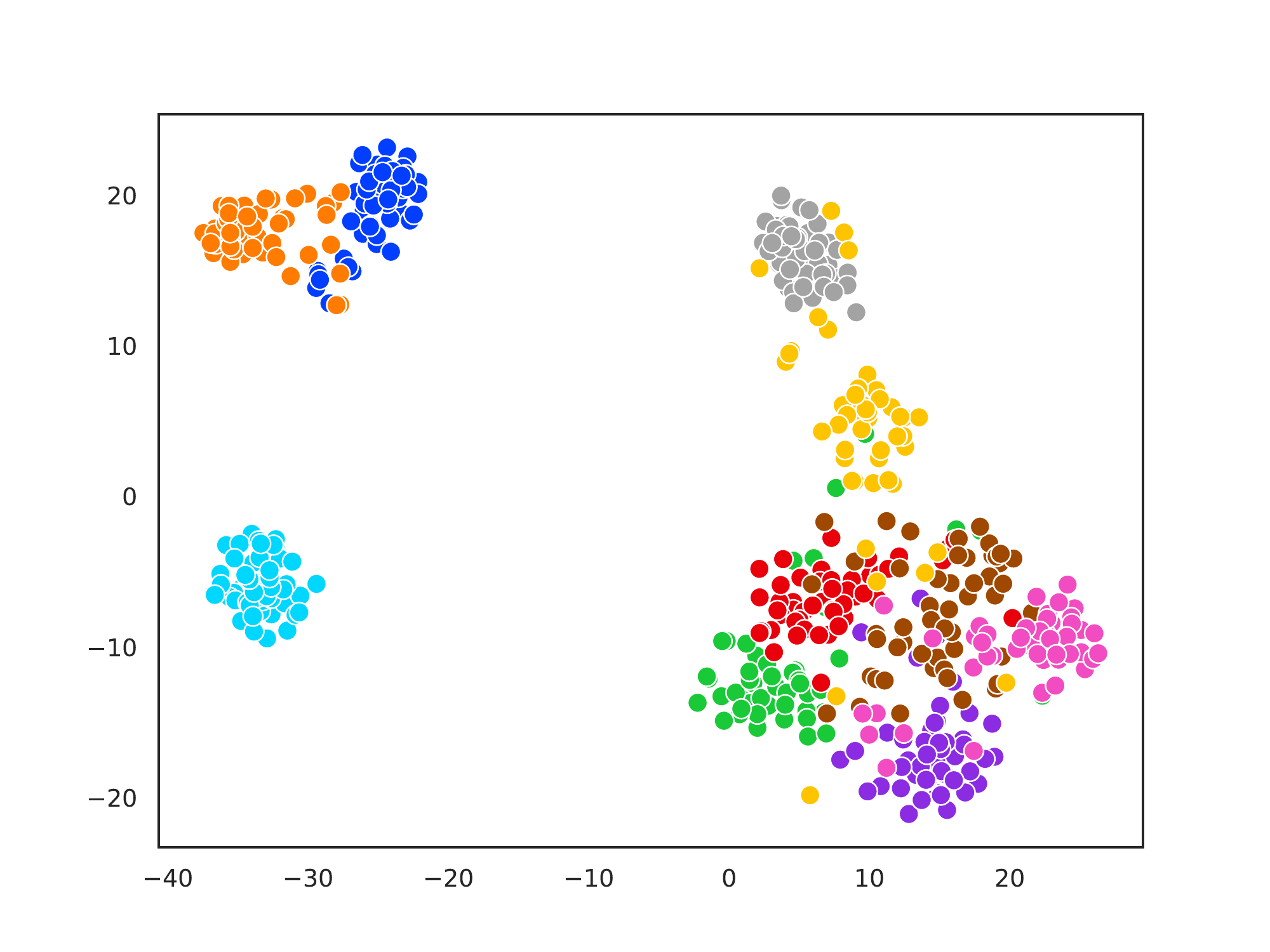}}\hfill
    %\subcaption{Weight distribution}
    \hspace{-8mm}
	\subcaptionbox{Joint training.}{\includegraphics[width = 1.0\linewidth]{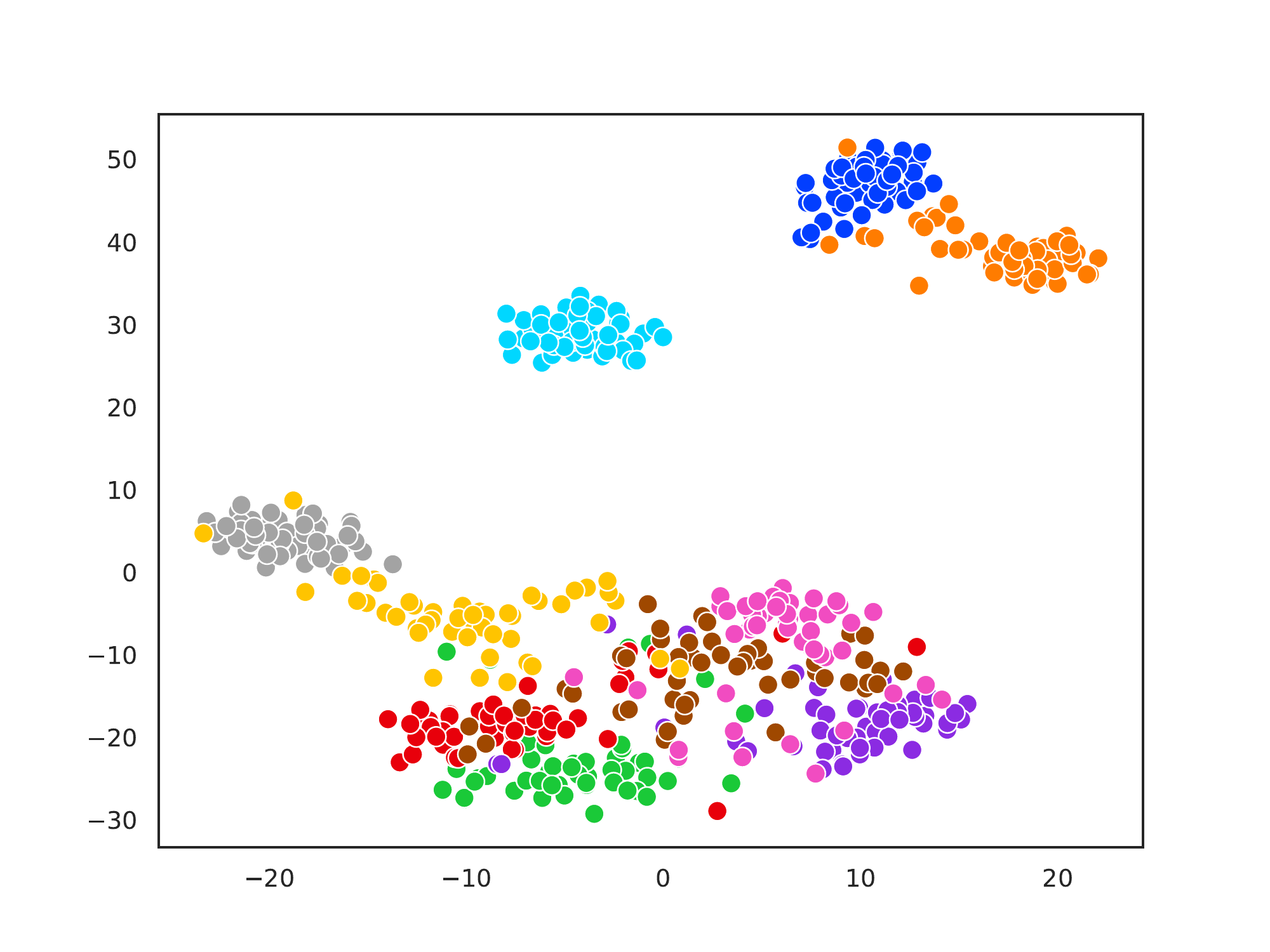}} 
    \hspace{-8mm}
	\subcaptionbox{Legends}{\includegraphics[width = 0.5\linewidth]{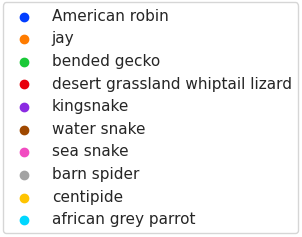}} 
	\caption{Comparisons of training vision-language model and linear adapter decoupled and jointly.}
	\vspace{-15pt}
	\label{ttsne}
\end{figure}

As discussed in Sec.~\ref{subsec:adapter} in main manuscript, decoupled training of vision-language model and linear adapter largely boost the performance, especially for few-shot categories. We visualize the classification space of several few-shot categories using t-SNE~\cite{van2008visualizing} as shown in Fig.~\ref{ttsne}. It is clearly illustrated in Sub-figure (a) that decoupled training achieves much more obvious separation boundary among different classes, especially for some easily confusing ones such as \textit{kingsnake} (purple), \textit{water snake} (brown) and \textit{sea snake} (pink).

% \begin{figure*}[ht!]
%   \centering
%   \vspace{-20pt}
%   \hspace{-2mm}
%     \subcaptionbox{Decoupled training.}{\includegraphics[width = 0.5\linewidth]{latex/images/decouple-tsne.pdf}}
%     %\subcaption{Weight distribution}
%     % \vspace{-1mm}
%     \hspace{-2mm}
% 	\subcaptionbox{Joint training.}{\includegraphics[width=0.5\linewidth]{latex/images/joint-tsne.pdf}}　
%     % \vspace{4mm}
% 	\subcaptionbox{Legend.}{\includegraphics[width = 0.3\linewidth]{latex/images/legend.png}}
% 	%\subcaption{Weight distribution}
% 	\caption{Comparisons of training vision-language model and linear adapter decoupled and jointly.}
% 	\vspace{-15pt}
% 	\label{tsne}
% \end{figure*}

\end{document}